\documentclass[lettersize,journal]{IEEEtran}
\usepackage{amssymb}
\usepackage{multirow}
\usepackage{algorithm}
\usepackage{algorithmic}
\usepackage{cite}

\usepackage{amsmath,amsfonts}
\usepackage{algorithmic}
\usepackage{array}
\usepackage[caption=false,font=normalsize,labelfont=sf,textfont=sf]{subfig}
\usepackage{textcomp}
\usepackage{stfloats}
\usepackage{url}
\usepackage{verbatim}
\usepackage{graphicx}
\def\BibTeX{{\rm B\kern-.05em{\sc i\kern-.025em b}\kern-.08em
    T\kern-.1667em\lower.7ex\hbox{E}\kern-.125emX}}
\usepackage{balance}
\usepackage{booktabs}
\usepackage{colortbl}
\usepackage{xcolor}
\usepackage{hyperref}

\begin{document}
\title{Diffusion Model Based Visual Compensation Guidance and Visual Difference Analysis for No-Reference Image Quality Assessment }
% \title{Diff$V^2$IQA: Diffusion Model and Vision transformer Based Network for No-Reference Image Quality Assessment}
% \author{IEEE Publication Technology Department
\author{Zhaoyang Wang, Bo Hu,  Mingyang Zhang, \textit{Member, IEEE,} Jie Li, Leida Li, \textit{Member, IEEE,} Maoguo Gong, \textit{Fellow, IEEE,} Xinbo Gao, \textit{Fellow, IEEE} 
\thanks{
This work was supported in part by the National Natural Science Foundation of China under Grants 62036007, 62101084, 62221005, 62171340; in part by the Natural Science Foundation of Chongqing under Grants CSTB2023NSCQ-LZX0085 and CSTB2023NSCQ-BHX0187; in part by the Creative Research Groups of Chongqing Municipal Education Commission under Grant CXQT21020; in part by the Science and Technology Research Program of Chongqing Municipal Education Commission under Grant KJQN202100628 and Grant KJQN202200638; and in part by the Chongqing Postdoctoral Research Project Special Support under Grant 2022CQBSHTB2052. (Corresponding authors: Xinbo Gao and Bo Hu.)

Zhaoyang Wang and Jie Li are with the State Key Laboratory of Integrated Services Networks, School of Electronic Engineering, Xidian University, Xi’an, Shaanxi 710071, China (e-mail: zywang23@stu.xidian.edu.cn; leejie@mail.xidian.edu.cn). 

Bo Hu is with the Chongqing Key Laboratory of Image Cognition, Chongqing University of Posts and Telecommunications, Chongqing 400065, China (emai: hubo90@cqupt.edu.cn)

Mingyang Zhang and Maoguo Gong are with the Key Laboratory of Collaborative Intelligent Systems, Ministry of Education, Xidian University, Xi’an 710071, China (e-mail: omegazhangmy@gmail.com; gong@ieee.org). 

Leida Li is with the School of Artificial Intelligence, Xidian University, Xi’an 710071, China (e-mail: ldli@xidian.edu.cn).

Xinbo Gao is with the School of Electronic Engineering, Xidian University, Xi’an 710071, China (e-mail: xbgao@mail.xidian.edu.cn), and with the Chongqing Key Laboratory of Image Cognition, Chongqing University of Posts and Telecommunications, Chongqing 400065, China (e-mail: gaoxb@cqupt.edu.cn).
}}

\markboth{Journal of \LaTeX\ Class Files,~Vol.~18, No.~9, September~2020}%
{How to Use the IEEEtran \LaTeX \ Templates}

\maketitle

\begin{abstract}
% Existing free-energy guided No-Reference Image Quality Assessment (NR-IQA) methods still suffer from finding a balance between learning feature information at the pixel level of the image and capturing high-level feature information
% and the efficient utilization of the obtained high-level feature information remains a challenge.
% As a novel class of state-of-the-art (SOTA) generative model, the diffusion model exhibits the capability to model intricate relationships, enabling a comprehensive understanding of images and possessing a better learning of both high-level and low-level visual features. Firstly, we devise a new diffusion restoration network that leverages the produced enhanced image and noise-containing images, incorporating nonlinear features obtained during the denoising process of the diffusion model, as high-level visual information. 
% Secondly, two visual evaluation branches are designed to comprehensively analyze the obtained high-level feature information. These include the visual compensation guidance branch, grounded in the transformer architecture and noise embedding strategy, and the visual difference analysis branch, built on the ResNet architecture and the residual transposed attention block. 

Existing free-energy guided No-Reference Image Quality Assessment (NR-IQA) methods continue to face challenges in effectively restoring complexly distorted images. The features guiding the main network for quality assessment lack interpretability, and efficiently leveraging high-level feature information remains a significant challenge.
As a novel class of state-of-the-art (SOTA) generative model, the diffusion model exhibits the capability to model intricate relationships,
enhancing image restoration effectiveness. Moreover, the intermediate variables in the denoising iteration process exhibit clearer and more interpretable meanings for high-level visual information guidance.
In view of these, we pioneer the exploration of the diffusion model into the domain of NR-IQA.
We design a novel diffusion model for enhancing images with various types of distortions, resulting in higher quality and more interpretable high-level visual information. Our experiments demonstrate that the diffusion model establishes a clear mapping relationship between image reconstruction and image quality scores, which the network learns to guide quality assessment. Finally, to fully leverage high-level visual information, we design two complementary visual branches to collaboratively perform quality evaluation.
Extensive experiments are conducted on seven public NR-IQA datasets, and the results demonstrate that the proposed model outperforms SOTA methods for NR-IQA. 
The codes will be available at \href{https://github.com/handsomewzy/DiffV2IQA}{https://github.com/handsomewzy/DiffV2IQA}.

\end{abstract}


\section{Introduction}
\IEEEPARstart{W}{ith} the rapid advancement of digital technologies, the growing significance of high-quality visual information has garnered widespread attention. To address the inevitable quality loss during the compression and transmission of visual data, image quality assessment (IQA) becomes crucial for ensuring the widespread dissemination of top-tier visual content. Categorized into subjective and objective IQA, the former relies on human perception, which is prone to biases and is resource-intensive. Objective IQA, particularly no-reference IQA (NR-IQA), emerges as a vital solution, employing computer algorithms for automatic image quality prediction. NR-IQA is particularly relevant in distorted scenarios without a reference image.

\begin{figure}[t]
    \centering
    \includegraphics[width=0.9\linewidth]{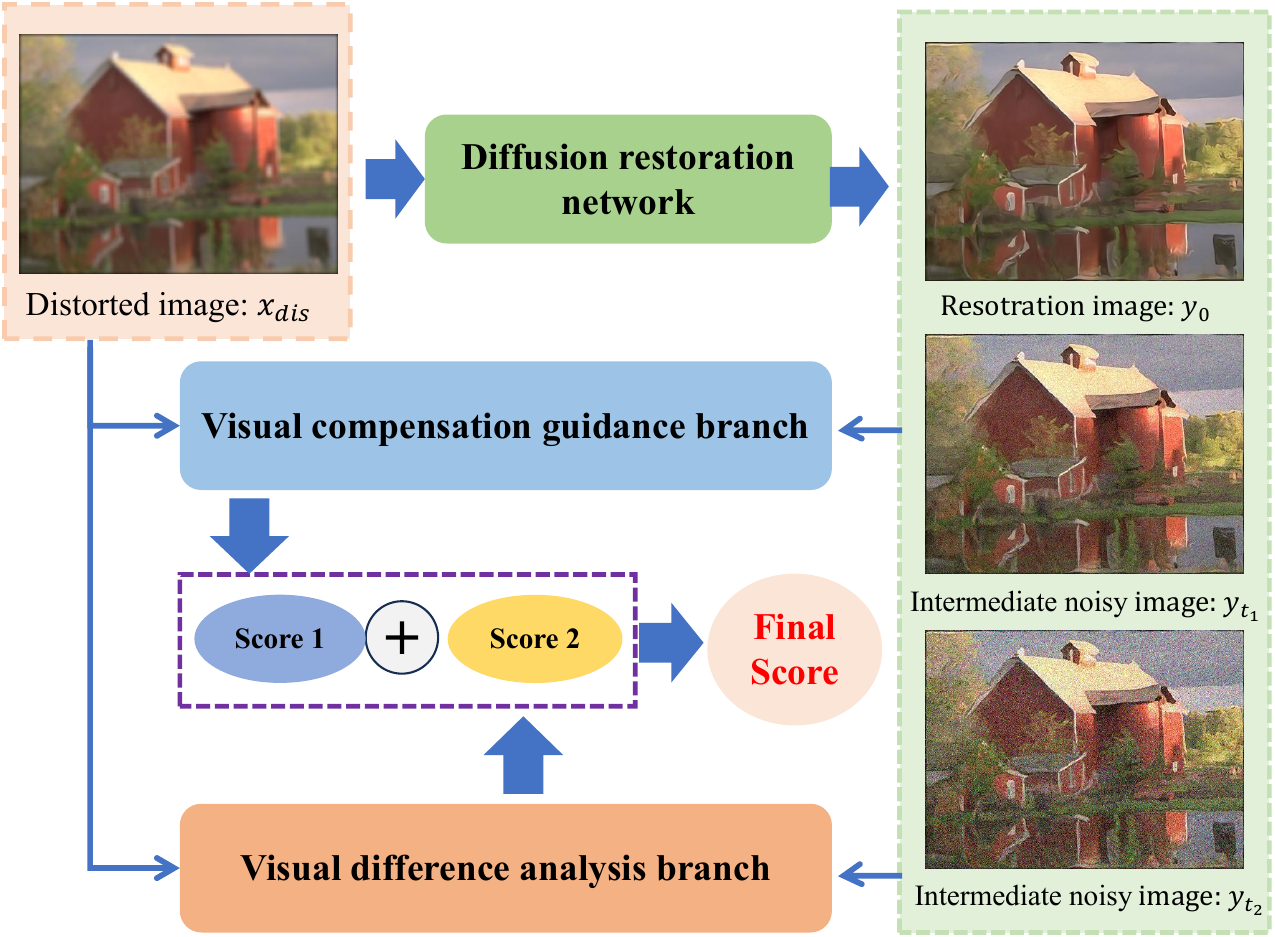}
    \caption{The overview framework of our proposed Diff$V^2$IQA model, which consists of two main components. The first part comprises a diffusion restoration network designed for image enhancement, while the second part involves a two-branch image quality evaluation network.  }
    \label{fig1}
    \vspace{-3mm}
\end{figure}

NR-IQA can be broadly categorized into two types based on the nature of image distortion: synthetic distortion and authentic distortion \cite{pan2022vcrnet}. In synthetic distortion NR-IQA, intentionally generated low-quality data incorporates various distortions such as JPEG artifacts, contrast distortions, and blurring \cite{sheikh2006statistical}. The resulting model primarily focuses on different distortion categories for the final evaluation and scoring.
In an authentic distortion dataset, low-scoring images are not associated with specific distortions but rather with high-level image information, such as slightly blurry characters or poor overall composition \cite{ghadiyaram2015massive, hosu2020koniq}. This introduces challenges, requiring the network to consider not only pixel-level image distortion but also prioritize high-level image information.

Early methods relied on hand-crafted features, broadly classified into two groups. One category \cite{fang2017no, fang2014no, li2016blind, moorthy2011blind, mittal2012no} utilizes static features of the image for quality assessment, while the other \cite{li2011blind,ye2012unsupervised,xu2016blind,xu2016multi,gu2017learning} employs machine learning methods to understand the distribution of specific image features, mapping them to the final quality score. However, these methods have limitations and struggle to adequately characterize image features, facing challenges in accurately describing image properties for various types of distortion.
With the advent of deep learning, especially convolutional neural networks (CNNs), some feature extraction methods based on CNN \cite{kang2014convolutional, zhang2018blind, su2020blindly, chen2021perceptual, zhou2021no, xu2022quality, jiang2017optimizing} have shown significant success in NR-IQA. These approaches leverage the inherent capability of convolution to autonomously extract crucial features from the image, significantly enhancing the precision of NR-IQA.
As deep learning advances, researchers explore the integration of successful architectures from various domains into NR-IQA, such as architectures based on the free-energy principle \cite{ren2018ran4iqa, lin2018hallucinated}, graphical convolutional neural networks (GNN) \cite{sun2022graphiqa}, transformer-based architectures \cite{ke2021musiq, golestaneh2022no, yang2022maniqa}, contrast learning principle \cite{zhao2023pretrain, madhusudana2022image}, test-time adaptation strategy \cite{roy2023test}, and large language models \cite{wang2023exploring, zhang2023blind}.
All these methodologies aim to improve the model's ability to comprehend and analyze the image for a more precise quality score.

Among the aforementioned methods, the approach based on the free energy principle, which autonomously enhances image quality akin to human observation, closely aligns with the process of image evaluation by humans. This method has garnered significant attention from scholars due to its promising research potential. 
In the latest free-energy guided study \cite{pan2022vcrnet}, the authors proposed the VCRNet network achieving optimal results; however, several shortcomings still need to be addressed: 
\begin{itemize}
\item \textbf{Model Architecture}: The performance of the image enhancement network utilized in the study could be further improved \cite{NEURIPS2020_4c5bcfec, pmlr-v139-nichol21a, li2022srdiff, saharia2022image}. Additionally, the main ResNet-based quality evaluation network lags behind existing Transformer architectures \cite{ke2021musiq, golestaneh2022no, yang2022maniqa}, which generally offer superior performance.
\item \textbf{Feature Interpretability}: The authors used multilayer variables in the restoration network U-Net to guide the quality evaluation network. However, it remains unclear whether there is a mapping between image quality and image reconstruction for these intermediate variables, and whether they effectively represent the self-repair function of the human eye in the context of the free energy principle. 
\item \textbf{Feature Utilization}: Only a subset of the available feature variables is effectively utilised, resulting in a significant portion of the data remaining underutilised.
\end{itemize}

To address these challenges, we propose a dual visual branch evaluation network based on the diffusion model (see Fig. \ref{fig1}). 
% This approach marks the first application of the diffusion model in NR-IQA, where it optimizes the architecture of the image restoration network enhancing its capability to effectively restore complexly distorted images, and significantly enhances feature interpretability, establishing a clear linear mapping relationship between image reconstruction and quality that can be learned by the network (see Fig. \ref{diff_vis}, Fig. \ref{vis_auth} and \textit{Sec. IV-B-4)-a), b)}). The dual-branching architecture combining Transformer and ResNet compensates for each other's strengths, greatly improving the ability to extract and utilize advanced visual information. 
This approach marks the first application of the diffusion model in NR-IQA, where it optimizes the architecture of the image restoration network enhancing its capability to effectively restore complexly distorted images.
Compared to the previous free-energy guided approach, our model provides high-level visual information that results in a more interpretable restorative image. We have quantitatively and qualitatively demonstrated that this approach aligns more closely with the self-repair function of the human eye (see \textit{Sec. IV-B-4)-a), b)}). Furthermore, unlike prior methods that use high-level visual information solely as a visual compensation mechanism, our approach incorporates a branching architecture for image disparity analysis, inspired by the FR-IQA concept, leveraging the enhanced interpretability of our feature information, which functions analogously to the additional `reference image'.

In summary, the main contributions of our work are as follows:
\begin{itemize}
    \item To the best of our knowledge, our work represents the first application of the diffusion model to the field of NR-IQA. Leveraging the free-energy principle, we devise a new diffusion image restoration model aimed at generating enhanced images as high-level visual features, demonstrating the potential of the diffusion model in NR-IQA.
    % which offers greater interpretability and better aligns with the self-repair function of the human eye.
    \item Our model demonstrates a clear enhancement in image quality, aligning closely with the free-energy principle that governs the human eye's self-repair mechanism. This alignment endows the model with a high degree of interpretability, bridging performance and theoretical insight.
    \item To fully exploit the rich high-level visual information provided by the diffusion model, we design two evaluation branches. The first branch focuses on the intrinsic features of the distorted image, using higher-level image information as visual compensation guidance. The second branch analyzes the difference between the distorted and restored images, capturing the nuanced changes.

    % \item The proposed `\textbf{Diff}usion model based \textbf{V}isual compensation guidance and \textbf{V}isual difference analysis' \textbf{IQA} network (Diff$V^2$IQA) has conducted extensive experiments on seven IQA databases, and achieved the SOTA performance in terms of prediction accuracy and monotonicity.
\end{itemize}

The remainder of this paper is structured as follows: In Section II, we provide an overview of related works on NR-IQA methods based on deep learning and discuss the development of the diffusion model. Section III provides an in-depth explanation of the proposed approach, while Section IV showcases the results obtained from experiments.
In Section V, we outline the limitations of the proposed model, potential avenues for improvement, and our future work directions.
Finally, Section VI serves as the concluding segment of this paper. 

\section{Related Works}

\subsection{No-Reference Image Quality Assessment}
% \textbf{Single-task CNN-based methods.}
In recent years, the rapid advancement of deep learning has given rise to CNN networks, leading to a proliferation of CNN-based NR-IQA \cite{kim2017deep}.
In \cite{kang2014convolutional}, Kang \textit{et al.} set a precedent by introducing CNN networks into the field of NR-IQA for the first time. 
Inspired by this work, many subsequent NR-IQA architectures based on CNN networks have emerged \cite{bosse2017deep, bianco2018use, yan2018two, po2019novel}.
During this period, the experimental datasets predominantly consisted of synthetic distorted datasets with a small data size. The above network fails to fully leverage the additional information about the type of data distortion and encounters challenges related to poor model generalization performance in the presence of limited training data. 
To tackle these challenges, several multi-task CNN frameworks have been proposed. 
In \cite{liu2017rankiqa}, Liu \textit{et al}. introduced RankIQA, which involves both dataset expansion and quality prediction tasks. This approach effectively addresses the challenge of limited IQA dataset size by employing a Siamese network.
% In \cite{ma2017end}, Ma \textit{et al}. presented MEON, a NR-IQA method based on a multi-task end-to-end deep neural network. MEON comprises two sub-networks, designed for distortion identification and quality prediction. 
In \cite{ma2017end}, Ma \textit{et al}. introduced MEON, a NR-IQA method utilizing a multi-task end-to-end deep neural network. MEON consists of two sub-networks designed for distortion identification and quality prediction.
Building on this foundation, in \cite{zhang2018blind}, Zhang \textit{et al}. proposed DBCNN, which employs the pre-trained CNN network (VGG16) as an initial feature extractor. They leverage the extracted higher-order features to perform distortion classification and quality prediction tasks, significantly enhancing accuracy. 
In \cite{su2020blindly}, Su \textit{et al}. proposed HyperIQA and they further improved the performance by using ResNet50 as a feature extractor and analysing data from different feature layers together.
However, with the advancement of dataset sizes and the inclusion of authentic distorted datasets, most of the CNN models of the above multitasking architectures share the same architecture, and performance bottlenecks emerge, particularly in capturing higher-level features and aesthetic features, and exhibiting an excessive focus on pixel-level evaluation when dealing with authentic distorted datasets. 

% \textbf{Restoration network guided methods.}
To improve the model's capacity to learn higher-level image information. 
Some researchers motivated by the free-energy principle observed in the human visual system, which inherently possesses the ability to automatically restore and improve distorted images. NR-IQA methods guided by the free-energy principle have been devised. These methods explore the quality reconstruction relationship between a distorted image and its restored counterpart. 
Ren \textit{et al.} introduced a restorative adversarial network for NR-IQA known as RAN4IQA \cite{ren2018ran4iqa}. Additionally, Lin \textit{et al.} proposed an NR-IQA method called Hallucinated-IQA \cite{lin2018hallucinated}, which is based on adversarial learning.
They leveraged the concept of GAN-based image generation to augment distorted images, providing guidance to the network for quality evaluation. 
In \cite{pan2022vcrnet}, Pan \textit{et al.} employed the U-Net network for image enhancement, addressing the challenges associated with GAN networks, including training difficulty and convergence issues. They also introduced a visual compensation module to support the network in learning higher-level enhancement information. However, there is room for further exploration of the image enhancement network's performance, and feature learning networks also require enhancement. 

% \textbf{Transformer-based methods.}
Furthermore, some researchers have incorporated the transformer architecture, known for its enhanced feature capabilities, into NR-IQA to augment the model's capacity for learning higher-level information. In \cite{ke2021musiq}, Ke \textit{et al.} introduced the MUSIQ architecture, showcasing the transformer architecture for the first time. 
In \cite{golestaneh2022no},  Alireza Golestaneh \textit{et al.} also utilize the transformer architecture and introduce relative ranking strategy and self-consistency strategy to improve the model performence.
In \cite{yang2022maniqa}, Yang \textit{et al.} extensively explored the efficiency of the transformer architecture in NR-IQA tasks, securing the first place in the NTIRE 2022 Perceptual Image Quality Assessment Challenge Track 2: No-Reference. 
The success of the transformer architecture in this challenge underscores its tremendous potential in NR-IQA, which also inspires us to use such architecture in our model.

Moreover, some researchers have incorporated other effective framework methods into NR-IQA. In \cite{zhao2023pretrain}, Zhao \textit{et al.} utilized the MoCo-V2 \cite{chen2020improved} framework and introduced the contrast learning method into NR-IQA, presenting the QPT model. However, it's worth noting that contrast learning exhibits a strong dependence on computational power. In \cite{roy2023test}, Roy \textit{et al.} introduced the test-time adaptation strategy to enhance the model's accuracy during testing across different test sets. In \cite{wang2023exploring}, Wang \textit{et al.} integrated the large language model (LLM) into NR-IQA, leveraging additional higher-level semantic information from the LLM to support the network in quality evaluation. 
Chen \textit{et al.} introduce TOPIQ in \cite{chen2023topiq}, presenting a top-down approach where high-level semantics guide the IQA network to emphasize semantically significant local distortion regions.

\subsection{Diffusion Model}
Most recently, the diffusion model \cite{NEURIPS2020_4c5bcfec, pmlr-v139-nichol21a} has emerged as a record-breaking model, demonstrating superior performance in generation and reconstruction tasks. This model has found widespread application in natural image restoration networks \cite{NEURIPS2020_4c5bcfec, pmlr-v139-nichol21a, li2022srdiff, saharia2022image}, particularly in the domain of super-resolution.
The image super-resolution task, which involves learning a mapping from a low-resolution image to a high-resolution image, shares a notable similarity with the mapping of a distortion type to a reference image in the context of IQA work. This similarity has inspired me to investigate the use of sophisticated diffusion model super-resolution architectures for the image restoration task within free-energy principle guided NR-IQA methods \cite{ren2018ran4iqa, lin2018hallucinated, pan2022vcrnet}.
It is worth mentioning that the SR3 model \cite{saharia2022image}, based on the diffusion model, has achieved high-performance results in super-resolving natural images. Additionally, the Stable Diffusion (LDM) \cite{rombach2022high} stands out as another top-performing diffusion method, exhibiting exceptional performance in super-resolution tasks.
Furthermore, the diffusion model has been integrated into the domain of FIQA. In \cite{babnik2023diffiqa}, Babnik \textit{et al.} introduced the DifFIQA model, marking the first application of the diffusion model in FIQA. They emphasize two crucial considerations: the balance between perturbation robustness and reconstruction quality. Specifically, high-quality images prove more resistant to perturbations while being easier to reconstruct. Based on this groundwork, the authors employ the diffusion model to restore face images and evaluate face quality. This assessment involves utilizing intermediate variables obtained from enhanced images within a FIQA framework. 
In this work, the training of the diffusion model necessitates reference face images, and it remains a challenge to effectively apply the diffusion model in the NR-IQA domain.

\begin{figure*}[t]
    \centering
    \includegraphics[width=0.9\linewidth]{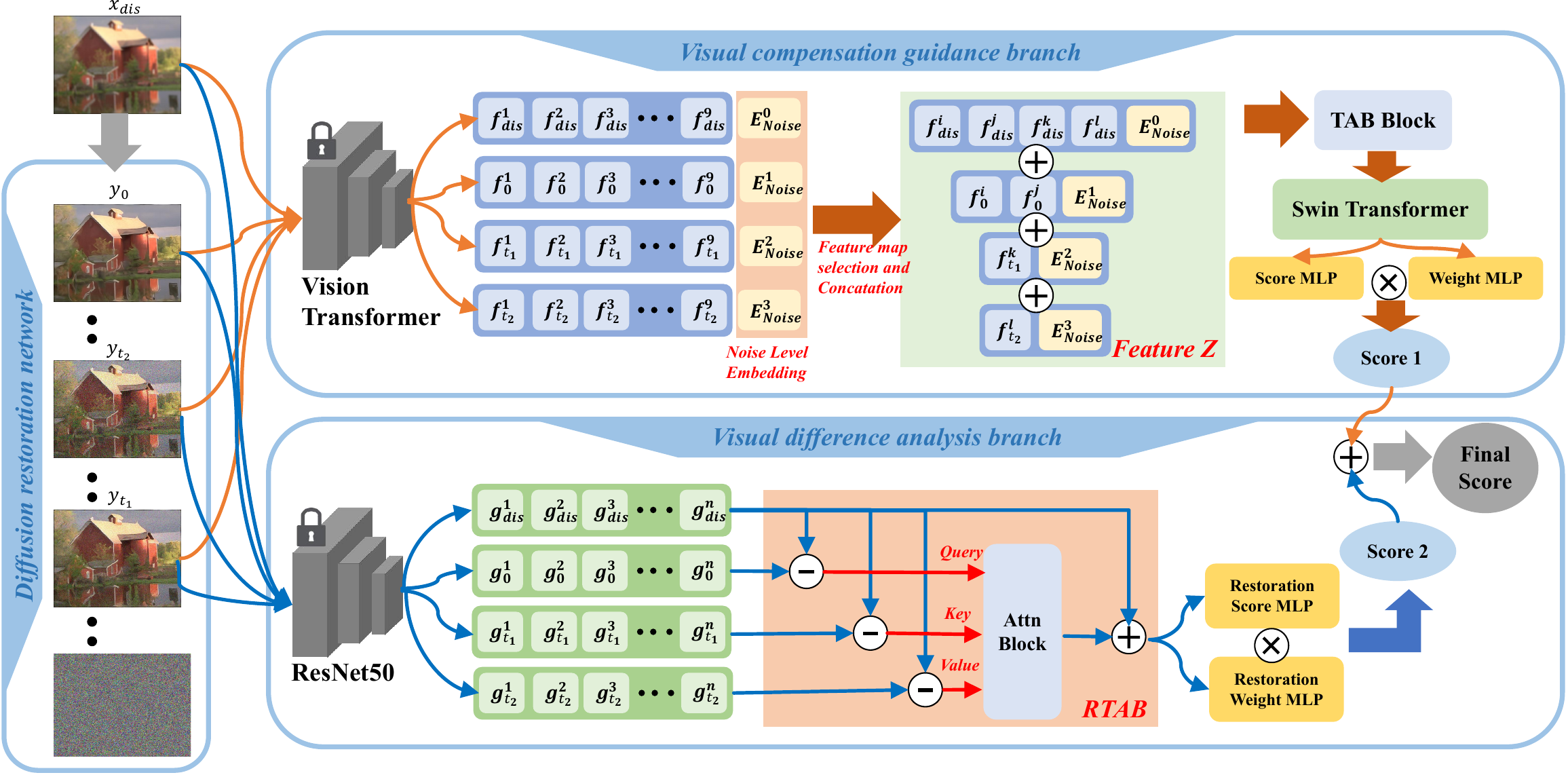}
    \caption{The illustration of proposed \textbf{Diff}usion model based \textbf{V}isual compensation guidance and \textbf{V}isual difference analysis' \textbf{IQA} network (Diff$V^2$IQA). We begin by feeding the input image into a diffusion restoration network, which generates the final restored image along with two intermediate noise-containing images. These images are then processed through two branches: the Visual Compensation Guidance (VCG) branch and the Visual Difference Analysis (VDA) branch. The VCG branch, built on a vision transformer architecture, employs a noise-level embedding strategy to produce a quality score. The VDA branch, based on ResNet50, incorporates an attention mechanism to generate its own score. The outputs from both branches are then combined and analyzed to yield the final quality score.} 
    \label{main}
% \vspace{-3mm}
\end{figure*}

% \subsection{Motivation}

\section{Proposed Diffusion Model Based Visual Compensation
Guidance and Visual Difference Analysis NR-IQA Method}
% Divided into four essential sub-sections, this chapter conducts a thorough examination of the overall architecture of our proposed model. The subsequent sections include three crucial components: the diffusion restoration network, visual compensation guidance branch, and visual difference analysis branch. Each sub-section intricately dissects the fundamental principles and functionalities of these components, offering a detailed comprehension of their roles within the overall framework.
% which lays the foundation for a profound exploration of our model's design and its effectiveness in addressing challenges in NR-IQA. 

\subsection{Overall Architecture }
The overall framework of our model (see Fig. \ref{main}.) is primarily composed of three key components: the diffusion restoration network, the visual compensation guidance branch based on vision transformer (ViT) \cite{dosovitskiy2020image}, and the visual difference analysis branch based on the residual attention module. Initially, the distorted image is input to the diffusion restoration network, generating one enhanced image and two noise-containing images. Following that, these resultant images, along with the original distorted image, are fed into the two visual evaluation branches, obtaining two respective scores, which are combined with designated weights to get the final quality score. 
In the visual compensation guidance branch, we use ViT for feature extraction, followed by the incorporation of noise level embedding strategy. Specific features are selected, spliced, and fused to capture both the original distorted image features and the high-level visual information introduced by the diffusion restoration network. Additionally, this process includes regularized information containing nonlinear noise. In the visual difference analysis branch, we initially employ ResNet50 for the preliminary encoding of visual features. Then all data is input into the RTAB for a holistic assessment of the differentiated information, ultimately yielding a score. In the following sections, we will provide a detailed exposition of the design principles and architectures underlying each branch module.

\begin{figure*}[t]
\centering
\includegraphics[width=0.9\textwidth]{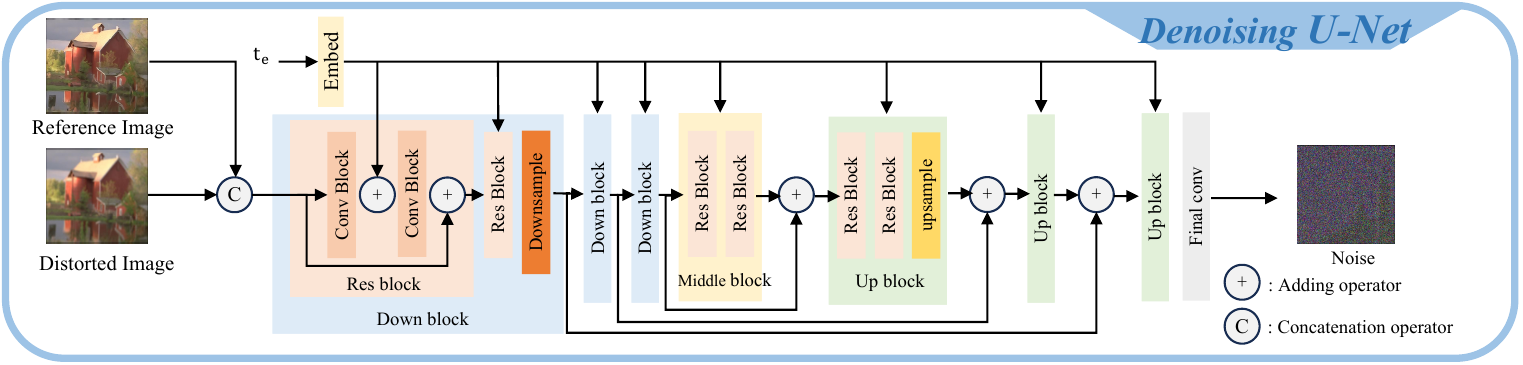} % Reduce the figure size so that it is slightly narrower than the column. Don't use precise values for figure width.This setup will avoid overfull boxes.
\caption{Overview of the pre-training process of the denoising model U-Net in our diffusion restoration network.}
\label{stage2}
% \vspace{-3mm}
\end{figure*}

\subsection{Diffusion Restoration Network }
\subsubsection{A Brief Introduction to Diffusion Model}
Diffusion models belong to the class of likelihood-based models, and they leverage the U-Net architecture to reconstruct
 the distribution of training data. The U-Net is trained to eliminate noise from training data deliberately distorted by the addition of Gaussian noise.
These models involve a forward noising process and a reverse denoising process. During the forward process, 
Gaussian noise is incrementally added to the original training data, denoted as $x_0 \sim q(x_o)$, in a step-by-step manner over T time steps,
following a Markovian process:
\begin{equation}
    q(x_t|x_{t-1}) = \mathcal{N} (\sqrt{1- {\beta}_t}x_{t-1}, {\beta}_tI)
\end{equation}
where $\mathcal{N} (\cdot)$ is a Gaussian distribution, the Gaussian variances ${\beta}_{t=0}^T$ that determine the noise schedule can be learned or scheduled. 
% Because of the preceding approach, an arbitrary noisy sample $x_t$ for each timestep $t$ is produced directly from $x_0$:
Due to the preceding method, each timestep $t$ generates an arbitrary noisy sample $x_t$ directly from the initial sample $x_0$:
\begin{equation}
    x_t = \sqrt{\bar{\alpha_t}}x_0 + \sqrt{1-\bar{\alpha_t}}\epsilon, \epsilon \sim \mathcal{N}(0,I)
\end{equation}
where $\alpha_t = 1 - \beta_t$, and $\bar{\alpha_t} = \prod_{s=1}^{T}\alpha_s $.
% In the reverse process, the diffusion model also follows a Markovian process to denoise the noisy sample $x_T$ back to $x_0$.
In the reverse process, the diffusion model similarly adheres to a Markovian process to restore the noisy sample $x_T$ back to $x_0$.
 This denoising is performed step by step. 
 In the case of large $T$ and small $\beta_t$, the reverse transition probability is approximated as a Gaussian distribution. 
This Gaussian distribution is predicted by a neural network that is trained to learn the reverse transitions, yielding the denoised sample $x_0$,
which can be formulated as:
\begin{equation}
    p_{\theta} (x_{t-1}|x_t) = \mathcal{N} (x_{t-1};\mu_{\theta}(x_t,t),\sigma_{\theta}(x_t,t))
\end{equation}
% where the reverse process is re-parameterized by estimating
% $\mu_{\theta}(x_t, t)$ and $\sigma_{\theta}(x_t, t)$. $\sigma_{\theta}(x_t, t)$ is set to $\sigma_t^2I$, where $\sigma_t^2I$ is not learned.
The re-parameterization of the reverse process involves estimating $\mu_{\theta}(x_t, t)$ and $\sigma_{\theta}(x_t, t)$. It is noteworthy that $\sigma_{\theta}(x_t, t)$ is fixed at $\sigma_t^2I$, and the values of $\sigma_t^2I$ are not subject to learning.
We then use U-Net to predict the one remaining item $\mu_{\theta}(x_t,t)$, and
the parameterization of $\mu_{\theta}(x_t,t)$ is obtained as follows:
\begin{equation}
    \mu_{\theta}(x_t,t) = \frac{1}{\sqrt{\alpha_t}}(x_t - \frac{1-\alpha_t}{\sqrt{1-\bar{\alpha_t}}}\epsilon_{\theta}(x_t,t))
\end{equation}
The U-Net model, represented as $\epsilon_{\theta}(x_t,t)$, is optimized by minimizing the following loss function:
\begin{equation}
    \mathcal{L}(\theta) = E_{t,x_0.\epsilon}[(\epsilon - \epsilon_\theta (\sqrt{\bar{\alpha_t}}x_0 + \sqrt{1-\bar{\alpha_t}\epsilon},t))^2]
\end{equation}

Furthermore, when additional information is incorporated as conditional guidance for the network to converge, the final loss function is then formulated as :
\begin{equation}
    \mathcal{L}(\theta) = E_{t,x_0.\epsilon}[(\epsilon - \epsilon_\theta (\sqrt{\bar{\alpha_t}}x_0 + \sqrt{1-\bar{\alpha_t}\epsilon},t,x_{cond}))^2]
\end{equation}

\subsubsection{The Architecture of Denoising Model U-Net}

\begin{algorithm}[t]
    \caption{U-Net training process}
    \label{alg_stage1}
    % \textbf{Input}: Distorted image: $x_{dis}$, Reference image: $x_{ref}$, time step: $t$.
    \begin{algorithmic}[1] %[1] enables line numbers
    \STATE \textbf{Repeat:}
     \STATE  \hspace{1em} $\epsilon \sim \mathcal{N}(0,I),  \alpha \sim p(\alpha)$ \\
   \STATE \hspace{1em} Take a gradient descent step on:
  \hspace{1em} \[\bigtriangledown_{\theta} (\epsilon - \epsilon_\theta (\sqrt{\bar{\alpha_t}}x_{dis} + \sqrt{1-\bar{\alpha_t}}\epsilon,t,x_{ref}))^2  \]
    \STATE \textbf{Until} converged
    \end{algorithmic}
    \end{algorithm}

Our U-Net architecture is based on SR3 \cite{saharia2022image}, a diffusion network designed for natural image super-resolution tasks (Fig. \ref{stage2}). Traditionally, in super-resolution tasks, networks learn a single type of distortion mapping by generating a low-resolution image through bicubic interpolation and pairing it with a high-resolution counterpart. However, in the IQA dataset, which contains various low-quality images, achieving generalized performance is challenging.
To address this, we initially reduce the image chunk size, allowing for a significant increase in the batch size from 8 to 256. We also replace the original SR3 network's GroupNorm, designed for a small batch size, with BatchNorm, better suited for the substantially increased batch size. Additionally, we remove specific preprocessing operations, such as bicubic upsampling before inputting the reference image into the SR3 network, driven by maintaining a consistent image size before and after this stage of processing.

These adjustments enhance the network's ability to learn more generalized features, enabling it to execute image restoration tasks more effectively across a diverse set of images. Our approach aims to foster a broader understanding of image characteristics within the network, ultimately improving its overall performance and efficacy in image restoration tasks on the IQA dataset.

The training of the U-Net begins by feeding the distorted image into the network. This process involves using the reference image from the IQA dataset as conditional information, which is concatenated with the input image (Fig. \ref{stage2}). The network undergoes training until convergence following the steps outlined in Algorithm \ref{alg_stage1}.
In the image recovery process (Algorithm \ref{alg_stage2}), at each time moment $t$, we continuously predict the noise, perform denoising and restoration operations, resulting in the enhanced image. Unlike SR3, our requirements extend beyond the final restored image; we also need intermediate variables representing the noise-containing restoration images at moments $t_1$ and $t_2$.
This approach draws inspiration from architectures such as VCRNet, HyperIQA, and U-Net, which leverage multi-level and multi-scale feature representations to enhance model robustness and diversify features. Following this paradigm, outputs from multiple stages are selected to provide high-level visual information. Ablation experiments are conducted to validate the effectiveness and utility of this strategy.

\begin{algorithm}[t]
    \caption{Iterative image restoration process}
    \label{alg_stage2}
    % \textbf{Input}: Distorted image: $x_{dis}$, time step: $t$.
    \begin{algorithmic}[1] %[1] enables line numbers
    \STATE $y_T \sim \mathcal{N}(0,I)$
    \STATE \textbf{for} $t = T,...,1$ \textbf{do}:
     \STATE \hspace{1em} $z \sim \mathcal{N}(0,I)$ if $t > 1$, else $z = 0$
     \STATE \hspace{1em} $y_{t-1} = \frac{1}{\sqrt{\alpha_t}}(y_t - \frac{1 - \alpha_t}{\sqrt{1-\gamma_t}} f_{\theta}(x_{dis}, y_t, \gamma_t) + \sqrt{1 - \alpha_t}z$
     \STATE \textbf{end for}
     \STATE \textbf{return} $y_0, y_{t_1}, y_{t_2}$ ($1 < t_1. t_2 < T$)
    \end{algorithmic}
    \end{algorithm}

\subsection{Visual Compensation Guidance Branch }

Through the diffusion restoration network, we obtain the final restored image along with intermediate noisy images ($y_0, y_{t_1}, y_{t_2}$). Then, we input these results into the ViT-based visual compensation guidance branch, simultaneously considering the original distorted image ($x_{dis}$). 
We structured our branch following the MANIQA \cite{yang2022maniqa} architecture. Similar to the approach outlined in that paper, we initially feed all the data into the ViT for preliminary extraction of feature information. The specific equation is shown below:
\begin{equation}
\begin{aligned}
     f_{dis}^1, f_{dis}^2 \dots f_{dis}^9 &= \textit{\textbf{E}}_{ViT}(x_{dis})\\
     f_{0}^1, f_{0}^2 \dots f_{0}^9 &= \textbf{\textit{E}}_{ViT}(y_{0})\\
     f_{t_1}^1, f_{t_1}^2 \dots f_{t_1}^9 &= \textbf{\textit{E}}_{ViT}(y_{t_1})\\
     f_{t_2}^1, f_{t_2}^2 \dots f_{t_2}^9 &= \textbf{\textit{E}}_{ViT}(y_{t_2})\\
     \label{vit_en}
\end{aligned}
\end{equation}
where $x_{dis}$ represents the input original image, $y_0$ signifies the ultimate recovered image produced by the diffusion model, while $y_{t_1}$ and  $y_{t_2}$ denote the intermediate noise-containing images acquired during the denoising process of the diffusion model. 
$f_i^j$ denotes the newly obtained post-encoding feature maps, where $i$ (ranging from 0 to 9) represents the feature maps of the 9 positions obtained, and $j$ (which can be $dis, 0, t_1, t_2$ represents the corresponding images.

Following that, we choose partial feature maps for subsequent processing to mitigate excessive redundancy in feature information, which is in alignment with MANIQA.
With the integration of the diffusion model, involving the restored image and the noise-containing images, 
we design a new reconstruction approach named noise level embedding strategy for obtaining new feature data map.
% In the VCRNet \cite{pan2022vcrnet}, the model incorporates a network for image restoration; however, it leverages the original image information extensively for quality evaluation, rather than relying solely on the high-level information introduced by the restoration network. 
% Consequently, we devise an image feature reconstruction strategy that primarily utilizes the original image information and use the diffusion introduced higher-level information as the visual compensation guidance.
We choose four feature maps from the original image obtained from ViT, followed by two feature maps from the final recovered image, and finally one feature map from each noise-containing image, which primarily utilizes the original image information and treat the diffusion introduced higher-level information as the visual compensation guidance
In this process, inspired by the positional embedding principle in ViT, we propose a new noise level embedding strategy into the network to differentiate among the original image, the restored image, and the noise-containing images, as depicted in Fig. \ref{main}. 
The specific equation is as follows:
\begin{equation}
\begin{aligned}
     z_{dis} &= [f_{dis}^6, f_{dis}^7 ,f_{dis}^8 ,f_{dis}^9] \oplus \textbf{\textit{E}}_{Noise Code}\\
     z_{0} &= [f_{0}^6, f_{0}^7] \oplus \textbf{\textit{E}}_{Noise Code}\\
     z_{t_1} &= [f_{t_1}^8] \oplus \textbf{\textit{E}}_{Noise Code}\\
     z_{t_2} &= [f_{t_2}^9] \oplus \textbf{\textit{E}}_{Noise Code}\\
     % &\textbf{\textit{E}}_{Noise Code} \sim \mathbb{R}^{1 \times 4 \times Embedding Dim}
\end{aligned}
\end{equation}
where $f_{i}^j$ denotes the ViT-encoded feature variable from Eq. \ref{vit_en}, $\textbf{\textit{E}}_{Noise Code}$ denotes the noise level encoding, and $z_i$ denotes the corresponding new fusion variable.
The `$\oplus$' symbol represents the concatenation of feature variables.
For the original image, we opt to select a larger number of feature maps, with the expectation that the network can concentrate on crucial original distorted features. Simultaneously, we introduce additional high-level features and non-linear feature information as visual compensation guidance to aid in network learning. 
Following this, the four feature sets are concatenated to construct a new feature map, which is input into the posterior network for learning and training to ultimately obtain a quality score. 
\begin{equation}
    % \textbf{\textit{Z}} = Concatenate[z_{dis}, z_{0}, z_{t_1}, z_{t_2}]
    \textbf{\textit{Z}} = z_{dis} \oplus z_{0} \oplus z_{t_1} \oplus z_{t_2}
\end{equation}

% The network design adheres to MANIQA's architecture, which has demonstrated considerable effectiveness. We have made minimal alterations to preserve its original efficacy. 
Upon acquiring the new feature map  ($\textbf{\textit{Z}}$), we traverse through the successive module depicted in Fig. \ref{main} to derive the ultimate score ($Score_1$) for the visual compensation guidance branch. 
The first ViT serves as a feature mapping encoder, while the parameters of the subsequent Swin Transformer are iteratively trained as the main component of the model to enhance the interactivity of the local information processed by the TAB block (see Fig. \ref{main}), ultimately outputting the evaluation scores.

\subsection{Visual Difference Analysis Branch}
Previous free-energy guidance models have faced challenges in fully utilizing high-level visual features. For example, in the VCRNet network, intermediate variables at each stage of the image enhancement process are only used as conditional inputs to guide the evaluation networks.
In contrast, our network explicitly leverages restored images as meaningful high-level visual compensation information. This approach establishes a more direct relationship between the input and restored images, as opposed to the indirect connection between feature variables and the input image.
The high correlation and similarity between the high-level visual information and the input image introduces a fundamentally different approach to feature processing and application. This allows us to analyze the difference information between images for auxiliary quality evaluation, inspired by the FR-IQA method. By utilizing the diffusion model, we artificially introduce the `reference images', a strategy that was previously unachievable. When intermediate features are used as high-level guidance information, such analysis is not feasible because the meaning, feature dimensions, and feature space of those variables are misaligned.

\begin{figure}
    \centering
    \includegraphics[width=0.95\linewidth]{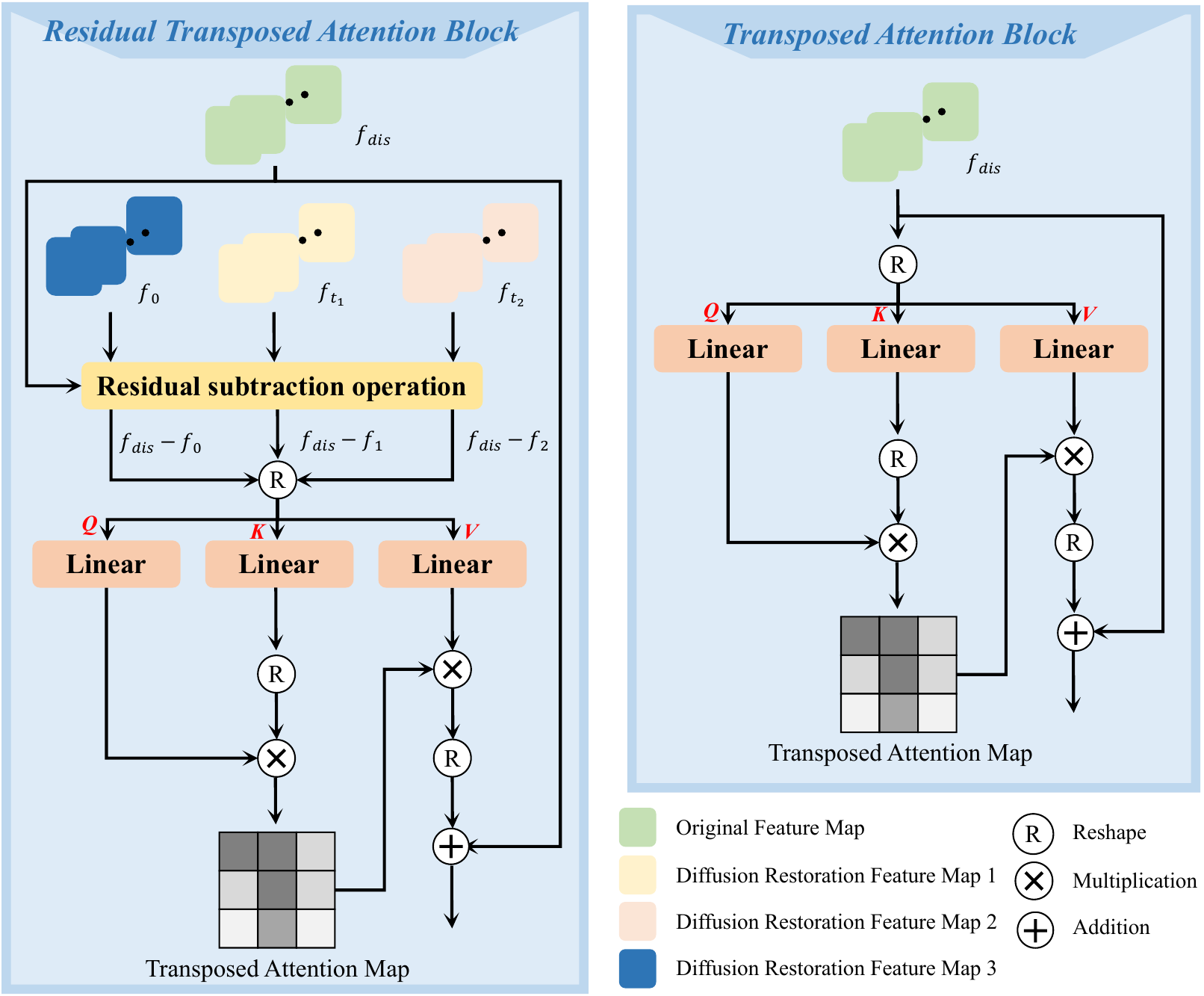}
    \caption{Comparison of our proposed Residual Transposed Attention Block with the Transposed Attention Block in the MANIQA paper and the specific architecture.}
    \label{rtab}
    \vspace{-3mm}
\end{figure}

First, all data are fed into ResNet50 for feature extraction. The extracted features are then passed into the RTAB module.
We deliberately chose not to continue using the transformer architecture for a specific reason: our focus is exclusively on capturing difference information. Using ViT could introduce an excessive number of high-level feature variables, unnecessarily complicating the model. Additionally, redirecting the data through ResNet into a new feature space enhances the model's overall robustness.
We structured our RTAB drawing inspiration from the transposed attention block (TAB) in MANIQA, and a comparative analysis along with specific differences is illustrated in Fig. \ref{rtab}. 
The TAB module incorporates a self-attention mechanism, utilizing itself as \textit{query} (Q), \textit{key} (K) and \textit{value} (V) for posterior network learning. In our adaptation, we consolidate the original image, the restored image, and the noise-containing images, performing a subtraction operation before inputting them into the subsequent network as Q, K, and V for learning, respectively. The network learns by emphasizing information about the disparities between the original image and the restored and noisy images. It concentrates on evaluating the difference between the distorted image and restored images, ultimately producing a score. 
The specific equation is shown below:
% \begin{equation}
% \begin{aligned}
%     &\textbf{Q} = |\textbf{\textit{E}}_{Res}(x_{\text{dis}}) - \textbf{\textit{E}}_{Res}(y_0)| \\
%     &\textbf{K} = |\textbf{\textit{E}}_{Res}(x_{\text{dis}}) - \textbf{\textit{E}}_{Res}(y_{t_1})| \\
%     &\textbf{V} = |\textbf{\textit{E}}_{Res}(x_{\text{dis}}) - \textbf{\textit{E}}_{Res}(y_{t_2})| \\
%     &\hat{\textbf{X}} = Attn(\textbf{\textbf{Q}, \textbf{K}, V}) + \textbf{X} \\
%     &Attn(\textbf{Q},\textbf{K},\textbf{V}) = \textbf{V} \cdot Softmax(\frac{\textbf{K} \cdot \textbf{Q}}{\alpha})
%     \label{attn}
% \end{aligned}
% \end{equation}
\begin{equation}
\begin{aligned}
    &\left\{
    \begin{aligned}
        &\textbf{Q} = |\textbf{\textit{E}}_{Res}(x_{\text{dis}}) - \textbf{\textit{E}}_{Res}(y_0)| \\
        &\textbf{K} = |\textbf{\textit{E}}_{Res}(x_{\text{dis}}) - \textbf{\textit{E}}_{Res}(y_{t_1})| \\
        &\textbf{V} = |\textbf{\textit{E}}_{Res}(x_{\text{dis}}) - \textbf{\textit{E}}_{Res}(y_{t_2})|
    \end{aligned}
    \right. \\
    &\hat{\textbf{X}} = Attn(\textbf{\textbf{Q}, \textbf{K}, V}) + \textbf{X} \\
    &Attn(\textbf{Q},\textbf{K},\textbf{V}) = \textbf{V} \cdot Softmax(\frac{\textbf{K} \cdot \textbf{Q}}{\alpha})
    \label{attn}
\end{aligned}
\end{equation}
% where $x_{dis}$ represents the input original image, $y_0$ signifies the ultimate recovered image produced by the diffusion model, while $y_{t_1}$ and  $y_{t_2}$ denote the intermediate noise-containing images acquired during the denoising process of the diffusion model. 

After the RTAB module, dual restoration MLP networks are employed to compute a score ($Score_2$) for the VDA branch. This design mirrors that of the VCG branch, utilizing adaptive weights provided by the weight MLP network. Subsequent ablation experiments have confirmed the robustness and effectiveness of this design.
The final score of our model is calculated as the sum of the scores from both branches.
% The weight $\alpha$ is set to 0.3 in practice.
\begin{equation}
    Score_{final} = Score_1 + Score_2
    \label{final_score}
\end{equation}

\section{Experiments}

\subsection{Experimental Settings}

\subsubsection{Datasets}
To ensure a rigorous and extensive evaluation of our proposed model's performance, our principal investigation incorporated seven diverse datasets. 

This ensemble consists of six synthetically distorted datasets: LIVE \cite{sheikh2006statistical}, CSIQ \cite{larson2010most}, TID2013 \cite{ponomarenko2015image}, Kadid10k \cite{lin2019kadid}, Waterloo Exploration Database \cite{ma2016waterloo} and PIPAL \cite{gu2020image},complemented by two authentic distortion datasets: the LIVE Challenge (CLIVE) \cite{ghadiyaram2015massive} and KonIQ10k \cite{hosu2020koniq}.  Experiments addressing individual distortion types are carried out on the synthetic distortion datasets, specifically LIVE and CSIQ. Additionally, we conduct cross-dataset validation experiments using the LIVE, CSIQ, TID2013, PIPAL and CLIVE datasets.
We conduct a study using the Waterloo Exploration Database and performed the group MAximum Differentiation (gMAD) competition game \cite{ma2016group} to assess the generalization ability of our model.
Finally, we investigate the pre-training datasets employed for our diffusion restoration network. We conduct comparative experiments to assess the impact of pre-training strategies based on the TID2013 and PIPAL datasets. 
% These experiments were designed to discern the variations and relative benefits of each pre-training approach within the context of our network's performance.

Consistent with established practices found in the majority of related research papers, we follow a standardized approach for processing datasets. 
Specifically, the training and testing datasets are randomly divided in an 8:2 ratio. During both training and testing phases, we employ random sampling of image blocks sized at $224 \times 224$. This choice is motivated by our use of the pretrained ViT model for feature extraction, necessitating conformity in data dimensions. 
% The $224 \times 224$ image block sampling ensures compatibility with the ViT model, enhancing the effectiveness of feature extraction during both training and testing. 
All reported results are obtained through 10 iterations of training and testing on the specific target dataset with randomly conducted splitting operations. The presented averages reflect the culmination of these results.

\subsubsection{Evaluation Metrics}
In our evaluation methodology, we incorporate two metrics that are extensively acknowledged for the assessment of model performance within the field. 
The first is Spearman’s rank order correlation coefficient (SRCC) and The second is Pearson’s linear correlation coefficient (PLCC).
% The first metric is Spearman’s rank order correlation coefficient (SRCC), which is computed as follows:
% \begin{equation}
%     \text{SRCC} = 1 - \frac{6\sum d_i^2}{n(n^2 - 1)}
% \end{equation}
% here, $d_i$ denotes the discrepancy between the rank-ordered predictions and the ground truth values, with $n$ representing the total sample count.
% The second metric is Pearson’s linear correlation coefficient (PLCC), defined by the equation:
% \begin{equation}
%     \text{PLCC} = \dfrac{\sum_{i=1}^n (x_i - \bar{x})(y_i - \bar{y})}{\sqrt{\sum_{i=1}^n (x_i - \bar{x})^2 \sum_{i=1}^n (y_i - \bar{y})^2}}
% \end{equation}
% where $x_i$ and $y_i$ correspond to the predicted and actual values respectively, and $\bar{x}$ and $\bar{y}$ are their mean values. 
Both SRCC and PLCC are bounded within the interval [-1, 1], with values approaching 1 signifying superior performance.

\subsubsection{Training Settings}
Our diffusion restoration network is pretrained on the PIPAL dataset, selected for its diverse range of distortion classes (116 categories). 
% In our ablation studies (see Table \ref{abl2}, Fig. \ref{diff_vis}), we also conduct comparative analyses to evaluate the impact of different pre-training datasets. 
For the diffusion restoration network, we use the Adam algorithm for optimization, with a learning rate of 1e-5. The restoration process is accelerated using a cosine schedule, and 50 diffusion time steps are uniformly set for both training and testing. To capture more extensive image features, we increase the batch size to 256, necessitating a corresponding reduction in the crop size to $16 \times 16$ pixels.
For the optimization of our image IQA network, we employ the Adam optimizer with a learning rate and weight decay factor set at 1e-5. Training utilizes the Mean Squared Error (MSE) loss function. Given the simultaneous processing of multiple image types, including original, restored, and noise-perturbed images, the batch size is set to 4 to accommodate computational demands.

\begin{table*}[t]
  \caption{Comparison of Diff$V^2$IQA vs. SOTA NR-IQA Algorithms on Six Standard Datasets (Best Two Results Highlighted in Bold and Underlined). The Results Are Sourced from Original Papers Such as MANIQA, Re-IQA, etc. A "-" Denotes Missing Results in the Original Papers. }
    \centering
    \small
% \resizebox{1\textwidth}{!}{
    \begin{tabular}{c|cccccccccccc}
    \toprule
         &  \multicolumn{2}{c}{LIVE}&  \multicolumn{2}{c}{CSIQ}&  \multicolumn{2}{c}{TID2013}&  \multicolumn{2}{c}{Kadid10k} &  \multicolumn{2}{c}{CLIVE}& \multicolumn{2}{c}{KonIQ10k}\\
         Method&  SRCC&  PLCC&  SRCC&  PLCC&  SRCC&  PLCC&  SRCC&  PLCC&  SRCC&PLCC& SRCC&PLCC\\
         \midrule
         BRISQUE\cite{mittal2012no} &  0.939&  0.935&  0.746&  0.829&  0.604&  0.694&  0.528&  0.567&  0.608&0.629& 0.665&0.681\\
         CORNIA\cite{ye2012unsupervised} &  0.947&  0.950&  0.678&  0.776&  0.678&  0.768&  0.516&  0.558&  0.629&0.671& 0.780&0.795\\
         DBCNN\cite{zhang2018blind}&  0.968&  0.971&  0.946&  0.959&  0.816&  0.865&  0.851&  0.856&  0.851&0.869& 0.875&0.884\\
         HyperIQA\cite{su2020blindly}&  0.962&  0.962&  0.923&  0.942&  0.840&  0.858&  0.852&  0.845&  0.859&0.882& 0.906&0.917\\
         CONTRIQUE\cite{madhusudana2022image}&  0.960&  0.961&  0.942&  0.955&  0.843&  0.857&  \underline{0.934}&  \underline{0.937}&  0.845&0.857& 0.894&0.906\\
 MUSIQ\cite{ke2021musiq}& -& -& -& -& -& -& -& -& -& -& 0.916&0.928\\
 GraphIQA\cite{sun2022graphiqa}& 0.979& 0.980& 0.947& 0.959& -& -& -& -& 0.845& 0.862& 0.911&0.915\\
 VCRNet\cite{pan2022vcrnet}& 0.973& 0.974& 0.943& 0.955& 0.846& 0.875& -& -& 0.856& 0.865& 0.894&0.909\\
 % QPT\cite{zhao2023pretrain}& -& -& -& -& -& -& -& -& \underline{0.895}& \textbf{0.914}& \textbf{0.927}&\textbf{0.941}\\
 CLIP-IQA\cite{wang2023exploring}& -& -& -& -& -& -& -& -& -& -& 0.895&0.909\\
 LIQE\cite{zhang2023blind}& 0.970& 0.951& 0.936& 0.939& -& -& 0.930& 0.931& \textbf{0.904}& \underline{0.910}& \underline{0.919}&0.908\\
         Re-IQA\cite{saha2023re}&  0.971&  0.972&  0.944&  0.964&  0.804&  0.861&  0.872&  0.885&  0.840&0.854& 0.914&0.923\\
 DEIQT\cite{qin2023data}& 0.980& 0.982& 0.946& 0.963& 0.892& 0.908& 0.889& 0.887& 0.875& 0.894& 0.921&0.934\\
 TReS\cite{golestaneh2022no}& 0.969& 0.968& 0.942& 0.922& 0.863& 0.883& 0.859& 0.858& 0.846& 0.877& 0.915&\underline{0.928}\\
         MANIQA\cite{yang2022maniqa}&  \underline{0.982}&  \underline{0.983}& \underline{ 0.961}&  \underline{0.968}&  \underline{0.937}& \underline{0.943}&  \underline{0.934}&  \underline{0.937}&  -&-& -&-\\
 TOPIQ\cite{chen2023topiq}& -& -& -& -& -& -& -& -& 0.870& 0.884& \textbf{0.926}&\textbf{0.939}\\
 \midrule
 \rowcolor{blue!5} % 整行背景色
 Diff$V^2$IQA(8:2)& \textbf{0.984}& \textbf{0.987}& \textbf{0.986}& \textbf{0.990}& \textbf{0.962}& \textbf{0.966}& \textbf{0.935}& \textbf{0.938}& \underline{0.897}& \textbf{0.913}& 0.917&\textbf{0.939}\\
 \rowcolor{blue!5} % 整行背景色
  Diff$V^2$IQA(7:1:2)& 0.975& 0.985& 0.979& 0.989& 0.953&0.956 & 0.928& 0.929& 0.851& 0.872& 0.918&0.938\\
  \rowcolor{blue!5} % 整行背景色
  Diff$V^2$IQA(6:2:2)& 0.973& 0.978& 0.975& 0.982& 0.946& 0.954& 0.927& 0.930& 0.837& 0.855& 0.917&0.936\\
     \bottomrule
    \end{tabular}
    \label{ex1}
    \vspace{-3mm}
\end{table*}

\begin{table*}[t]
    \centering
\caption{SRCC Results for Individual Distortion Types on the LIVE and CSIQ Databases. The Results Are Sourced from Original Papers such as VCRNet and GraphIQA. (Best Two Results Highlighted in Bold and Underlined)}
\label{individual}
    \small
    \begin{tabular}{c|ccccc|cccccc}
    \toprule
         &  \multicolumn{5}{c|}{LIVE}&  \multicolumn{6}{c}{CSIQ}\\
         Method&  JP2K&  JPEG&  WN&  GBLUR&  FF&  JP2K&  JPEG&  WN&  GBLUR& FN&CC\\
    \midrule
         DIIVINE\cite{moorthy2011blind}&  0.925&  0.913&  0.985&  0.958&  0.845&  0.803&  0.732&  0.756&  0.785& 0.432&0.789\\
         BRISQUE\cite{mittal2012no}&  0.914&  0.965&  0.979&  0.951&  0.877&  0.840&  0.806&  0.723&  0.820& 0.378&0.804\\
         CORNIA\cite{ye2012unsupervised}&  0.936&  0.934&  0.962&  0.926&  0.912&  0.831&  0.513&  0.664&  0.836& 0.493&0.462\\
         HOSA\cite{xu2016blind}&  0.928&  0.936&  0.964&  0.964&  0.934&  0.818&  0.733&  0.604&  0.841& 0.500&0.716\\
 DeepIQA\cite{bosse2017deep}& 0.968& 0.953& 0.979& 0.970& 0.897& 0.934& 0.922& 0.944& 0.901& 0.867&0.847\\
 TSCNN\cite{yan2018two}& 0.959& 0.965& 0.981& 0.970& 0.930& 0.922& 0.902& 0.923& 0.896& 0.873&0.866\\
 DBCNN\cite{zhang2018blind}& 0.963& 0.976& 0.982& 0.971& 0.918& 0.946& 0.951& 0.947& 0.939& 0.943&0.883\\
 HyperIQA\cite{su2020blindly}& 0.949& 0.961& 0.982& 0.926& 0.936& 0.960& 0.934& 0.927& 0.915& 0.931&0.874\\
 RAN4IQA\cite{ren2018ran4iqa}& 0.962& 0.922& 0.984& 0.972& 0.920& 0.936& 0.914& 0.931& 0.892& 0.835&0.859\\
 Hall-IQA\cite{lin2018hallucinated}& 0.969& 0.975& 0.992& 0.973& 0.953& 0.924& 0.933& 0.942& 0.901& 0.842&0.861\\
 GraphIQA \cite{sun2022graphiqa}& \textbf{0.979}& \underline{0.978}& 0.978& \underline{0.978}& \textbf{0.979}& 0.947& 0.947& \underline{0.948}& 0.947& \underline{0.948}&\underline{0.947}\\
 VCRNet\cite{pan2022vcrnet}& 0.975& \textbf{0.979}& \underline{0.988}& \underline{0.978}& 0.962& \underline{0.962}& \underline{0.956}& 0.939& \underline{0.950}& 0.899&0.919\\
         Diff$V^2$IQA&  \underline{0.977}&  \underline{0.978}&  \textbf{0.994}&  \textbf{0.991}&  \underline{0.970}&  \textbf{0.986}&  \textbf{0.973}&  \textbf{0.978}&  \textbf{0.972}& \textbf{0.988}&\textbf{0.958}\\
 \bottomrule
    \end{tabular}
    \vspace{-3mm}
\end{table*}

\subsection{Experimental Results}
\subsubsection{Performance Evaluation on Individual Database}

To evaluate the robustness and effectiveness of our proposed model, we conduct a comprehensive set of experiments involving six datasets—four synthetic distortion datasets and two authentic distortion datasets. We compare a total of 15 methodologies, including traditional approaches like BRISQUE \cite{mittal2012no} and CORNIA \cite{ye2012unsupervised}, as well as contemporary deep learning models. Among the deep learning-based contenders, we assess CNN-based architectures such as DBCNN \cite{zhang2018blind} and HyperIQA \cite{su2020blindly}, graph convolutional networks like GraphIQA \cite{sun2022graphiqa}, transformer-based models like MUSIQ \cite{ke2021musiq}, TReS \cite{golestaneh2022no}, and MANIQA \cite{yang2022maniqa}. We also include a model based on contrastive learning, CONTRIQUE \cite{madhusudana2022image}. We carefully select representative models from each technological domain to ensure a thorough and balanced comparison. 
Our training process strictly adheres to the experimental setups outlined in the source publications. Most reported results are directly sourced from the original papers. 
Most studies, including MANIQA, DEIQT, and TReS, have used an 8:2 experimental ratio, while others, such as VCRNet, Re-IQA, and LIQE, have employed a 7:1:2 split. To more fairly and efficiently demonstrate the validity of our method, we conduct experimental validation using various dataset divisions, as shown in Table \ref{ex1}. The table highlights three commonly used experimental splits with a blue background.
% Throughout the training, we ensured consistency in random number seeds, varying only in network architectures to ensure a high level of experimental fairness. 
% The results of these experiments are systematically documented in Table \ref{ex1}.

In Table \ref{ex1}, our newly developed Diff$V^2$IQA model consistently outperforms existing alternatives across all synthetic distortion datasets, including LIVE, CSIQ, TID2013, and Kadid10k, regardless of dataset size. Particularly noteworthy is the performance on the LIVE dataset, which is relatively straightforward. In this case, MANIQA shows commendable performance, making the improvements offered by our model somewhat modest. However, when dealing with the slightly more complex CSIQ and TID2013 datasets, our model's advancements in metric performance are notably evident, achieving optimal scores of 0.986 and 0.990 on CSIQ, and 0.961 and 0.962 on TID2013, respectively. 
Our method achieves the best results across different experimental ratio settings. This is particularly evident when comparing it to the three methods: VCRNet, Re-IQA, and LIQE under the 7:1:2 ratio, as shown in Table \ref{ex1}.
Furthermore, our method demonstrates satisfactory performance on authentic distortion datasets. However, it does not achieve the optimal results. This is primarily due to the fact that our diffusion model's image enhancement on authentic distortion datasets is less effective compared to synthetic distortion datasets (see Fig. \ref{diff_vis}, Fig. \ref{vis_auth} and \textit{4) Visualisation Results: a), b)}).
This limits the level of visual guidance it can provide. Furthermore, it can be observed that the discrepancy in results is more pronounced when the training set changes, particularly in the case of smaller datasets compared to synthetic distortion datasets. This highlights the model's increased dependence on data.
In summary, our proposed model showcases commendable performance on both synthetic and authentic distortion datasets. 
% By incorporating high-level information through a diffusion restoration network, introducing non-linearity via noisy intermediate images, and subsequently employing a dual visual branch to discern the decoupling of diverse data types and establish a reconstructive link between distorted images and their respective quality scores, our approach achieves a high level of accuracy in quality assessment.

\begin{table}[t]
    \centering
\caption{SRCC Result of Cross Database Test. The Results Are Sourced from Original Papers. (Best Two Results Highlighted in Bold and Underlined)}
\label{crossdataset1}
\small
\resizebox{0.48\textwidth}{!}{
    \begin{tabular}{c|ccc|ccc}
    \toprule
         Training&  \multicolumn{3}{c|}{TID2013}&  \multicolumn{3}{c}{CLIVE}\\
         Testing&  LIVE&  CSIQ&  CLIVE&  LIVE&  TID2013& CSIQ\\
         \midrule
         BRISQUE\cite{mittal2012no}&  0.724&  0.568&  0.109&  0.244&  0.275& 0.236\\
         CORNIA\cite{ye2012unsupervised}&  0.809&  0.659&  0.263&  0.553&  0.385& 0.433\\
         CNN\cite{kang2014convolutional}&  0.532&  0.598&  0.104&  0.103&  0.021& 0.095\\
         DeepIQA\cite{bosse2017deep}&  0.805&  0.683&  0.009&  0.323&  0.141& 0.323\\
         DBCNN\cite{zhang2018blind}&  \underline{0.872}&  0.703&  \textbf{0.412}&  \textbf{0.757}&  \underline{0.401}& \textbf{0.631}\\
         RAN4IQA\cite{ren2018ran4iqa}&  0.811&  0.694&  0.036&  0.296&  0.157& 0.276\\
         VCRNet\cite{pan2022vcrnet}&  0.822&  \underline{0.721}&  0.307&  \underline{0.746}&  \textbf{0.416}& 0.566\\
         Diff$V^2$IQA&  \textbf{0.907}&  \textbf{0.829}&  \underline{0.340}&  0.708&  0.345& \underline{0.595}\\
         \bottomrule
    \end{tabular}
    }
    \vspace{-3mm}

\end{table}

\begin{table}[t]
    \centering
        \caption{Evaluations on Cross Datasets. Each model is trained using PIPAL and subsequently tested on the LIVE and TID2013 datasets. The Comparative Experimental Results Are Sourced from MANIQA. }
    \label{crossdataset2}
    \small
    \resizebox{0.46\textwidth}{!}{
    \begin{tabular}{cc|cccc}
    \toprule
         \multicolumn{2}{c}{Training on}&  \multicolumn{4}{c}{PIPAL}\\
         \midrule
          \multicolumn{2}{c}{Testing on}&  \multicolumn{2}{c}{LIVE}&  \multicolumn{2}{c}{TID2013}\\
         \multicolumn{2}{c}{}&  PLCC&  SRCC&  PLCC& SRCC\\
         \midrule
         \multirow{3}{*}{FR}&  PSNR&  0.873&  0.865&  0.687& 0.677\\
         &  WaDIQaM\cite{bosse2017deep}&  0.883&  0.837&  0.698& 0.741\\
         &  RADN\cite{shi2021region}&  0.905&  0.878&  0.747& 0.796\\
         \midrule
         \multirow{3}{*}{NR}&  TReS\cite{golestaneh2022no}&  0.643&  0.663&  0.516& 0.563\\
         &  MANIQA\cite{yang2022maniqa}&  0.835&  0.855&  0.704& 0.619\\
         &  Diff$V^2$IQA&  0.769&  0.805&  0.729& 0.652\\
    \bottomrule
    \end{tabular}
    }
    \vspace{-3mm}
\end{table}

\begin{figure*}[t]
    \centering
    \includegraphics[width=0.7\linewidth]{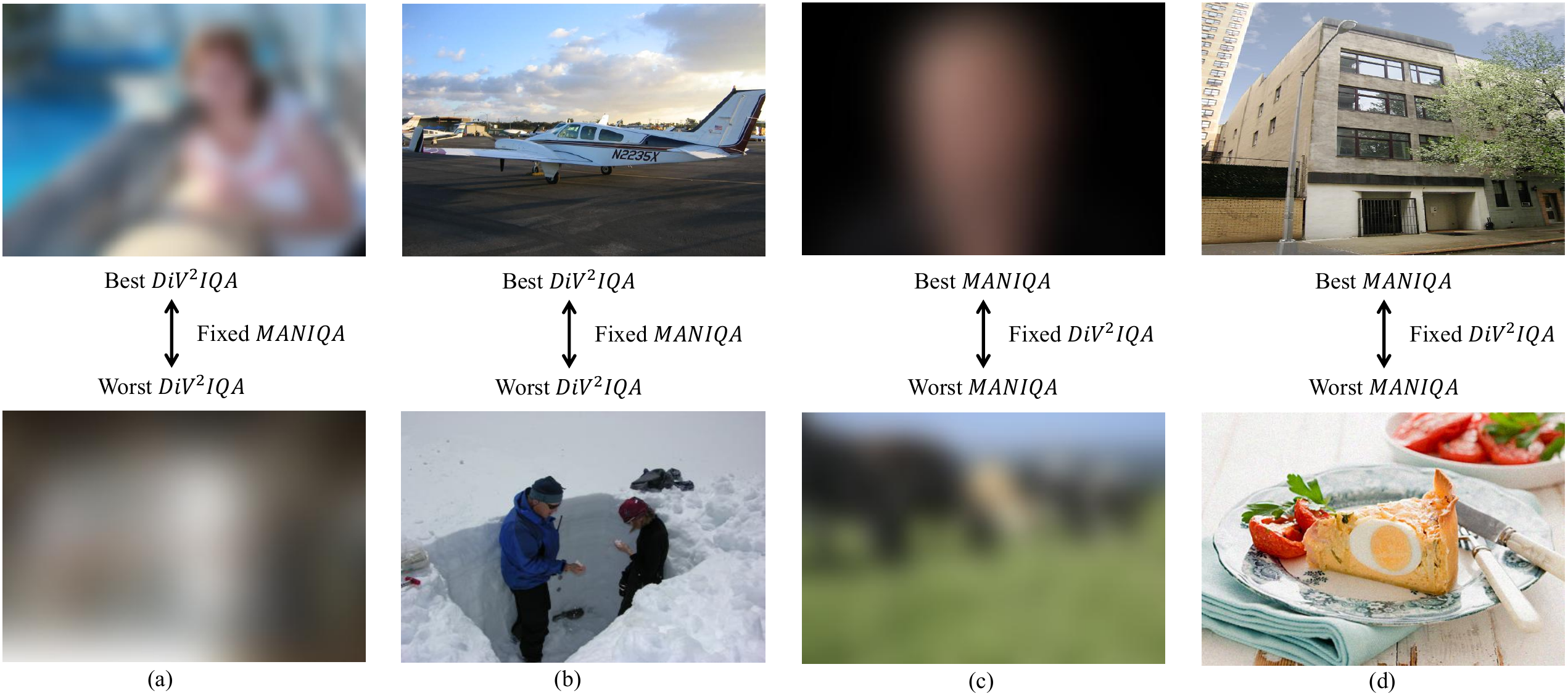}
    \caption{gMAD competition results between Diff$V^2$IQA and MANIQA. (a) Fixed MANIQA at the low-quality level. (b) Fixed MANIQA at the high-quality
level. (c) Fixed Diff$V^2$IQA at the low-quality level. (d) Fixed Diff$V^2$IQA at the high-quality level.}
    \label{M_D}
\vspace{-3mm}
\end{figure*}

\begin{figure*}[t]
    \centering
    \includegraphics[width=0.7\linewidth]{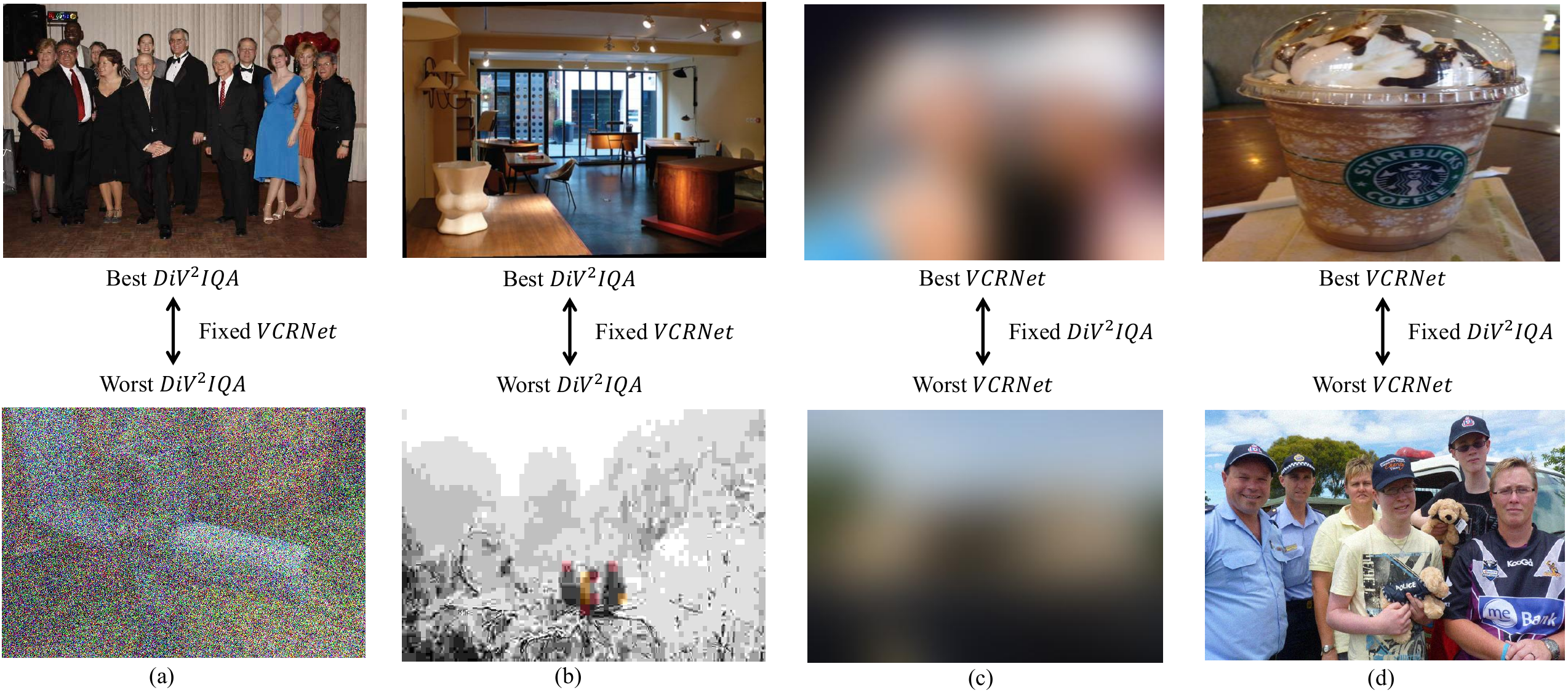}
    \caption{gMAD competition results between Diff$V^2$IQA and VCRNet. (a) Fixed VCRNet at the low-quality level. (b) Fixed VCRNet at the high-quality
level. (c) Fixed Diff$V^2$IQA at the low-quality level. (d) Fixed Diff$V^2$IQA at the high-quality level. }
    \label{V_D}
\vspace{-3mm}
\end{figure*}

% \begin{figure*}[t]
%     \centering
%     \includegraphics[width=0.8\linewidth]{visual_imgs.pdf}
%     \caption{Visualize the image restoration results on the TID2013 dataset for the diffusion restoration network. From top to bottom, four different types of distortions are addressed: Change of color saturation, Gaussian blur, Image color quantization with dither, and High frequency noise. In each row, column `a' represents the reference image, columns `b' to `e' show the distorted levels 1 and their corresponding final restored and noisy images for this type of distortion. Columns `f' to `i' represent distorted levels 5 and their corresponding final restored and noisy images. Beneath each image, there is an error map comparing the original image with the reference image. }
%     \label{diff_vis}
% \vspace{-3mm}
% \end{figure*}

\subsubsection{Performance Evaluation on Individual Distortion Types}
To substantiate the efficacy of our model, we extend our experimental framework to include analyses of individual distortion types, the results of which are delineated in Table \ref{individual}. 
Due to the fact that some methods' networks do not specifically focus on this aspect, experiments targeting on individual database were not conducted. Therefore, we selected representative methods that are more focused and specialized in this experimental segment for comparison.
Our investigations predominantly focuse on the LIVE and CSIQ datasets. We draw comparisons across twelve methods.
It is worth noting that the selection of methods compared during our individual distortion type experiments diverges from those included in earlier full dataset evaluations. This discrepancy arises due to the absence of individual distortion type experiment comparisons within the original publications of some previously examined methods, resulting in a lack of available comparative data.

Examination of the Table  \ref{individual} reveals that our proposed Diff$V^2$IQA model attains superior performance on the LIVE dataset, securing the highest scores in all categories except for Fast Fading (FF), including JPEG2000 (JP2K), JPEG, White Noise (WN), and Gaussian Blur (GBLUR) distortions. In the case of the CSIQ dataset, our model consistently delivers the best results across all evaluated distortion categories.
The aforementioned findings, in conjunction with prior experimental outcomes, mutually reinforce the conclusion that our model exhibits exceptional efficacy when applied to synthetically distorted datasets. This performance is largely attributable to the incorporation of high-level feature information facilitated by the diffusion restoration model, in synergy with the ViT model's robust feature learning capabilities. 
% The additional experiments focusing on singular distortion classes further substantiate the effectiveness of our proposed model.

\subsubsection{Performance Evaluation on Waterloo Exploration Database}

To assess the generalization performance of the model, gMAD experiments are conducted using the Waterloo Exploration Database. For comparison, two methods closely aligned with our approach—VCRNet and MANIQA—are selected, as depicted in Fig. \ref{M_D} and Fig. \ref{V_D}. The gMAD experiment is structured to evaluate a model’s limitations by attempting to induce failure; a model that resists failure is deemed robust. The experiment consists of two key components: the attacker and the defender, each fulfilling distinct roles in challenging and defending the model. The models are trained independently on the LIVE dataset and subsequently tested on the Waterloo Exploration Database, comprising 99,624 images.

% \textcolor{blue}{
The results of our model, in comparison to MANIQA, are presented in Fig. \ref{M_D}. In column (a), our model differentiates between image pairs that MANIQA judged as having the same low quality, emphasizing sharper contours to highlight both advantages and shortcomings. In column (b), where MANIQA rated both images as optimal, our model further distinguishes the superior images based on factors such as compositional layout, brighter colors, and overall perception closer to human visual preferences. When our model serves as the defender, the images show minimal differences.
As illustrated in Fig. \ref{V_D}, our model outperforms VCRNet in generalization performance. When acting as an attacker, our model identifies instances where VCRNet's judgments deviate from human vision. As a defender, it remains consistent with human visual assessments, showcasing stronger generalization overall.

\subsubsection{Performance Evaluation Cross Different Databases}

% In order to establish the generalization capability of the proposed Diff$V^2$IQA, we conduct experiments wherein the synthetically distorted IQA database was evaluated on the authentically distorted IQA database, and vice versa. 
To further assess the generalization capability of the proposed Diff$V^2$IQA, we perform experiments in which the IQA database synthetically distorted is evaluated on the authentically distorted IQA database, and vice versa.
The specific SRCC results are presented in Table \ref{crossdataset1}. 
Given the different setups in various papers, we chose the cross-dataset experiments from VCRNet and MANIQA, which are most relevant to our work, striving to cover as many datasets as possible to demonstrate the effectiveness of our model.
It is evident that all of our models consistently achieve top-2 metrics, even outperforming other models in several datasets. This substantiates the robust generalization ability of our proposed model. 
Additionally, given that the distortion types in synthetically and authentically distorted IQA databases are entirely different, it is challenging for IQA methods to generalize to authentically distorted databases when trained on synthetically distorted databases, and vice versa. 
In this scenario, our model continues to demonstrate excellent performance metrics, underscoring its robust generalization capabilities. While the diffusion restoration network of our model is trained on synthetic distortion, its effectiveness extends to a certain degree in handling authentic distorted images as well. 

Furthermore, to further substantiate the generalization ability of our model, we performed additional experiments following the cross-dataset approach outlined in the MANIQA paper, as depicted in Table \ref{crossdataset2}. The training is conducted on the PIPAL dataset, and testing is carried out on TID2013 and the LIVE dataset. 
A comprehensive analysis is conducted by evaluating both SRCC and PLCC results, which are compared against several FR-IQA methods to thoroughly demonstrate the performance of our model. Notably, the results on TID2013 are particularly noteworthy (PLCC: 0.727, SRCC: 0.649), even when compared to certain full-reference methods. This highlights the strong generalization performance of our model. In conclusion, the proposed Diff$V^2$IQA framework exhibits robust generalization capabilities for both synthetic and authentic distortions.

\begin{table*}[t]
 \caption{Image quality evaluations on the synthetic dataset using three NR-IQA methods for the diffusion model generated images.
 `a,b,c,d’ represent each of the four distortion types in Fig. \ref{diff_vis}.
 The percentages that follow in the data represent the percentage improvement from the distorted image, with red representing an improvement and blue representing a decrease.}
    \centering
% \small
\resizebox{0.9\textwidth}{!}{
    \begin{tabular}{cc|ccccccccc}
    \toprule
  & && \multicolumn{4}{c}{ Distorted level 1}& \multicolumn{4}{c}{Distorted level 5}\\
          &Metric &$x_{ref}$&  $x_{dis}$&  $y_0$&  $y_{t_1}$&  $y_{t_2}$&  $x_{dis}$& $y_0$& $y_{t_1}$&$y_{t_2}$\\
     \midrule
           &NIQE $\downarrow$ & 3.65 & 3.72 & 6.07/\textcolor{blue}{63.17\%} & 13.6/\textcolor{blue}{265.59\%} & 16.14/\textcolor{blue}{333.87\%} & 3.65 & 3.51/\textcolor{red}{3.84\%} & 13.4/\textcolor{blue}{267.12\%} & 14.17/\textcolor{blue}{288.22\%} \\
 A&MUSIQ $\uparrow$&77.66& 77.61 & 62.22/\textcolor{blue}{19.79\%} & 32.07/\textcolor{blue}{58.73\%} & 31.67/\textcolor{blue}{59.23\%} & 75.77 & 75.16/\textcolor{blue}{0.81\%} & 47.06/\textcolor{blue}{37.98\%} & 36.32/\textcolor{blue}{52.03\%}\\
  &TReS $\uparrow$&94.35& 93.10 & 57.70/\textcolor{blue}{38.04\%} & 17.37/\textcolor{blue}{81.35\%} & 15.76/\textcolor{blue}{83.09\%} & 85.63 & 85.88/\textcolor{red}{0.29\%} & 21.40/\textcolor{blue}{75.10\%} & 18.90/\textcolor{blue}{77.85\%}\\
  \midrule
 & NIQE $\downarrow$ & 3.14 & 2.98 & 3.9/\textcolor{blue}{30.87\%} & 9.92/\textcolor{blue}{232.89\%} & 13.53/\textcolor{blue}{354.03\%} & 8.46 & 5.58/\textcolor{red}{34.04\%} & 11.83/\textcolor{blue}{39.83\%} & 15.59/\textcolor{blue}{84.28\%} \\
 B&MUSIQ $\uparrow$
&73.38& 70.78 & 61.13/\textcolor{blue}{13.67\%} & 47.25/\textcolor{blue}{33.35\%} & 36.32/\textcolor{blue}{48.36\%} & 17.96 & 43.42/\textcolor{red}{141.13\%} & 27.41/\textcolor{red}{52.27\%} & 21.15/\textcolor{red}{17.56\%}\\
   &TReS $\uparrow$&88.39& 80.66 & 68.57/\textcolor{blue}{14.95\%} & 31.58/\textcolor{blue}{60.87\%} & 24.62/\textcolor{blue}{69.57\%} & 22.77 & 35.98/\textcolor{red}{58.02\%} & 20.71/\textcolor{blue}{9.05\%} & 21.01/\textcolor{blue}{7.75\%}\\
   \midrule
 & NIQE $\downarrow$ & 3.46 & 5.06 & 4.27/\textcolor{red}{15.60\%} & 11.00/\textcolor{blue}{117.39\%} & 13.08/\textcolor{blue}{158.30\%} & 11.88 & 4.00/\textcolor{red}{66.32\%} & 10.83/\textcolor{red}{8.86\%} & 13.68/\textcolor{blue}{15.14\%} \\
 C&MUSIQ $\uparrow$
&74.69& 72.08 & 63.86/\textcolor{blue}{11.39\%} & 44.96/\textcolor{blue}{37.75\%} & 36.51/\textcolor{blue}{49.33\%} & 51.01 & 54.63/\textcolor{red}{7.22\%} & 32.80/\textcolor{blue}{35.56\%} & 29.22/\textcolor{blue}{42.70\%}\\
  &TReS $\uparrow$&94.71& 90.35 & 68.89/\textcolor{blue}{23.73\%} & 21.47/\textcolor{blue}{76.26\%} & 17.03/\textcolor{blue}{81.12\%} & 72.67 & 65.56/\textcolor{blue}{9.79\%} & 23.67/\textcolor{blue}{67.47\%} & 18.35/\textcolor{blue}{74.74\%}\\
  \midrule
 & NIQE $\downarrow$ & 4.19 & 6.35 & 5.82/\textcolor{red}{8.35\%} & 12.06/\textcolor{blue}{90.08\%} & 15.80/\textcolor{blue}{148.43\%} & 18.86 & 4.85/\textcolor{red}{74.28\%} & 12.52/\textcolor{red}{33.61\%} & 13.95/\textcolor{red}{26.06\%} \\
 D&MUSIQ $\uparrow$
&76.04& 71.63 & 65.29/\textcolor{blue}{8.85\%} & 40.60/\textcolor{blue}{43.39\%} & 29.07/\textcolor{blue}{59.38\%} & 38.05 & 63.85/\textcolor{red}{68.18\%} & 37.53/\textcolor{blue}{1.37\%} & 28.83/\textcolor{blue}{24.14\%}\\
          &TReS $\uparrow$&92.03&  60.74 & 62.25/\textcolor{red}{2.49\%} & 18.03/\textcolor{blue}{70.35\%} & 16.26/\textcolor{blue}{73.25\%} & 19.72 & 61.37/\textcolor{red}{211.40\%} & 17.52/\textcolor{blue}{11.16\%} & 16.81/\textcolor{blue}{14.72\%}\\
    \bottomrule
    \end{tabular}
}
   
    \label{linear1}
    \vspace{-3mm}
\end{table*}

\begin{figure}[t]
    \centering
    \includegraphics[width=0.9\linewidth]{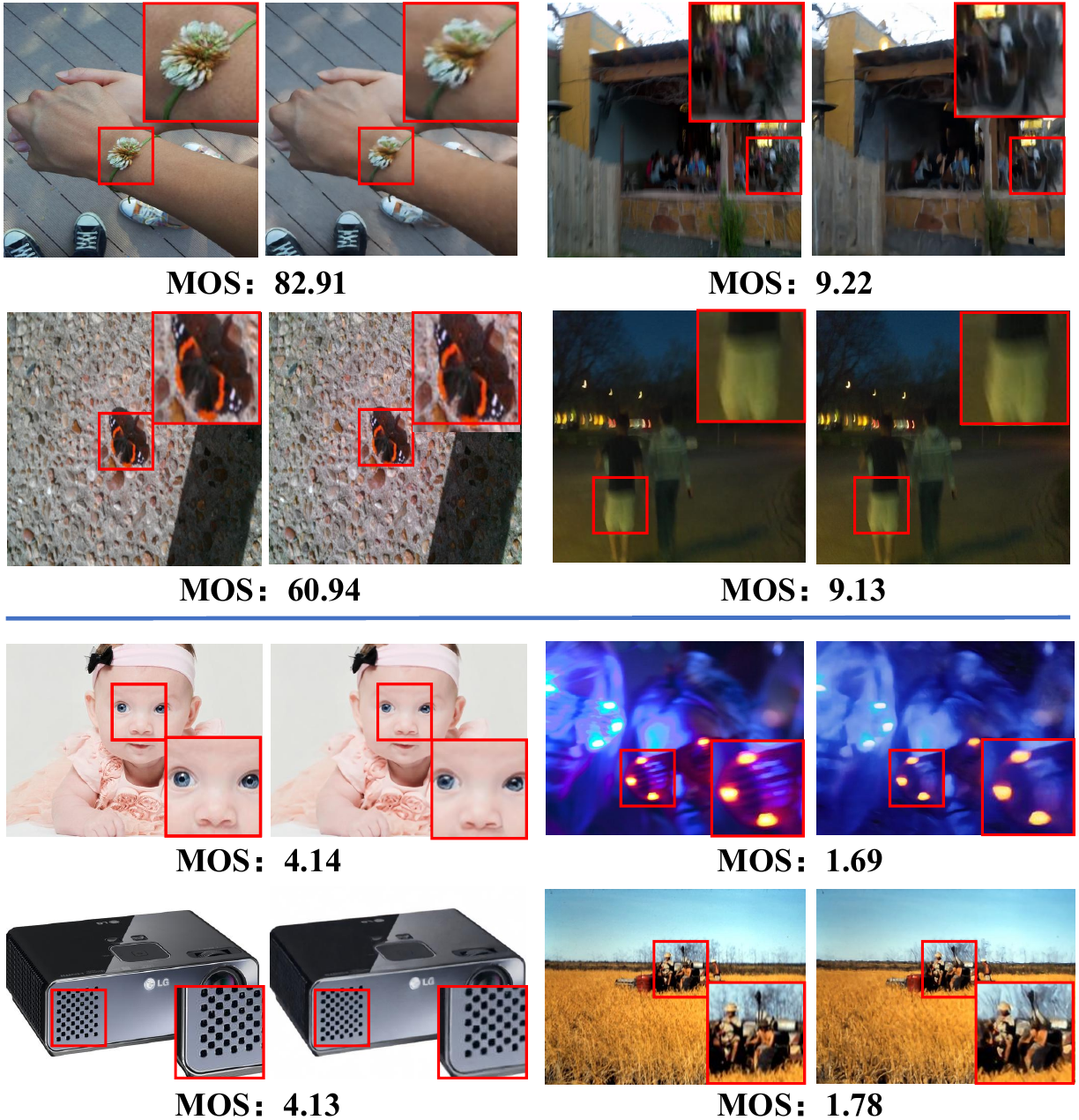}
    \caption{Visualization of the recovery results of the diffusion model on the authentic distortion IQA dataset. Each set of images shows the original and final restored images, respectively. The first two rows are from the CLIVE dataset, and the last two rows are from the KonIQ-10K dataset.}
    \label{vis_auth}
\vspace{-3mm}
\end{figure}

\begin{table}[t]
 \caption{Image quality evaluations on the authentic dataset using three NR-IQA methods for the diffusion model generated images. The percentages that follow in the data represent the percentage improvement from the distorted image, with red representing an improvement and blue representing a decrease.}
    \centering
\small
\resizebox{0.45\textwidth}{!}{
    \begin{tabular}{c|cccc}
    \toprule
   &  \multicolumn{2}{c}{ High MOS}&\multicolumn{2}{c}{Low MOS}\\
          Metric &  $x_{ori}$&  $y_0$&  $x_{ori}$&  $y_0$\\
     \midrule
           NIQE $\downarrow$ & 5.23 & 7.44/\textcolor{blue}{42.26\%} & 5.82 & 5.48/\textcolor{red}{5.84\%}\\
 MUSIQ $\uparrow$& 77.26 & 61.47/\textcolor{blue}{20.54\%} & 27.60 & 34.58/\textcolor{red}{25.20\%}\\
  TReS $\uparrow$& 105.88 & 96.46/\textcolor{blue}{8.87\%} & 32.25 & 37.30/\textcolor{red}{15.79\%}\\
  \midrule
  NIQE $\downarrow$ & 3.70 & 5.56/\textcolor{blue}{50.81\%} & 5.89 & 7.17/\textcolor{blue}{21.75\%}\\
 MUSIQ $\uparrow$
& 77.75 & 75.98/\textcolor{blue}{2.90\%} & 22.30 & 25.65/\textcolor{red}{15.11\%}
\\
   TReS $\uparrow$& 96.50 & 74.33/\textcolor{blue}{22.94\%} & 14.56 & 16.23/\textcolor{red}{11.52\%}
\\
   \midrule
  NIQE $\downarrow$ &5.18 & 5.20/\textcolor{blue}{0.39\%} & 4.71 & 4.42/\textcolor{red}{6.15\%}\\
 MUSIQ $\uparrow$
& 45.06 & 50.07/\textcolor{red}{11.18\%} & 54.75 & 52.46/\textcolor{blue}{4.18\%}
\\
  TReS $\uparrow$& 58.93 & 64.02/\textcolor{red}{8.56\%} & 37.04 & 57.02/\textcolor{red}{54.13\%}
\\
  \midrule
  NIQE $\downarrow$ & 5.27 & 6.26/\textcolor{blue}{18.71\%} & 7.76 & 6.66/\textcolor{red}{14.21\%}\\
 MUSIQ $\uparrow$
& 76.40 & 72.30/\textcolor{blue}{5.37\%} & 20.34 & 21.13/\textcolor{red}{3.89\%}
\\
          TReS $\uparrow$&  109.44 & 77.48/\textcolor{blue}{29.23\%} & 9.89 & 10.93/\textcolor{red}{10.55\%}
\\
    \bottomrule
    \end{tabular}
}
   
    \label{linear2}
    \vspace{-3mm}
\end{table}

\subsubsection{Visualisation Results}

\paragraph{\textbf{The visualization of the diffusion restored images on authentic distortions}}

In Fig. \ref{vis_auth}, we visualize the recovery results of the diffusion restoration network on  authentic distortion datasets. The first two rows show the visualization results on the CLIVE dataset, and the last two rows on the KonIQ-10k dataset, where we comparatively display the restoration results on authentic distorted images with different evaluation scores. Since the diffusion model is pre-trained on the synthetic distortion dataset PIPAL, achieving quality recovery results on the authentic distortion dataset, which presents more complex scenarios, is challenging. However, there are still some common features between synthetic and authentic distortions.
As illustrated in Fig. \ref{vis_auth}, for high-quality images (i.e., those with high MOS scores), the restoration results are nearly indistinguishable from the original images, and the model encounters challenges in restoring authentic distortions. In contrast, for lower-quality images (i.e., those with low MOS scores), the model applies more pronounced restorations, such as color correction in the third row of Fig. \ref{vis_auth}, and character contour restoration in the first and fourth rows.

Given that the human eye may struggle to precisely assess the efficacy of these restorations, three NR-IQA metrics are employed to objectively evaluate and compare the images before and after restoration, as shown in Table \ref{linear2}. 
In addition, to further validate the mapping relationship between image quality and reconstruction quality, we conduct additional tests on the synthetic distortion dataset. These tests included various distortion types as well as the quality assessment of intermediate noisy images, as presented in Table \ref{linear1}.
The results demonstrate a significant improvement in image quality for lower-quality images. 
The mapping relationship, where the model achieves greater quality improvements for images with poorer initial quality, is consistent across both authentic and synthetic distortion datasets.

The mapping relationship between image quality and reconstruction quality is a fundamental aspect of our model's architecture. Compared to VCRNet, our model operates with greater transparency, as both qualitative and quantitative evaluations demonstrate that high-level visual guidance images are optimized more effectively. These images exhibit a closer alignment with the natural self-repair process of the human visual system.

% \textcolor{blue}{
% Overall, the diffusion restoration model demonstrates a clear and explicit linear relationship between reconstruction quality and evaluation scores across both synthetic and authentic distortion datasets. This relationship serves as the core criterion guiding the network for quality evaluation.
% When confronted with synthetic distortion datasets, the diffusion model approximates the distribution of the images to match the quality level of the reference image, achieving better final restoration as image quality improves. Conversely, when dealing with authentic distortion datasets, the diffusion model shows limited improvement in processing high-quality, high-scoring images, maintaining overall picture quality in a "preserved" state. However, when processing severely distorted, low-scoring images, the significant difference between the distortion types and the generalized features learned previously by the model results in the final recovered image being in a "corrupted" state.
% Both scenarios exhibit a clear linear relationship between reconstruction quality and evaluation scores, which can be learned by the network, allowing it to utilize the feature information to improve overall performance.
% }

\begin{figure}[t]
    \centering
    \includegraphics[width=0.9\linewidth]{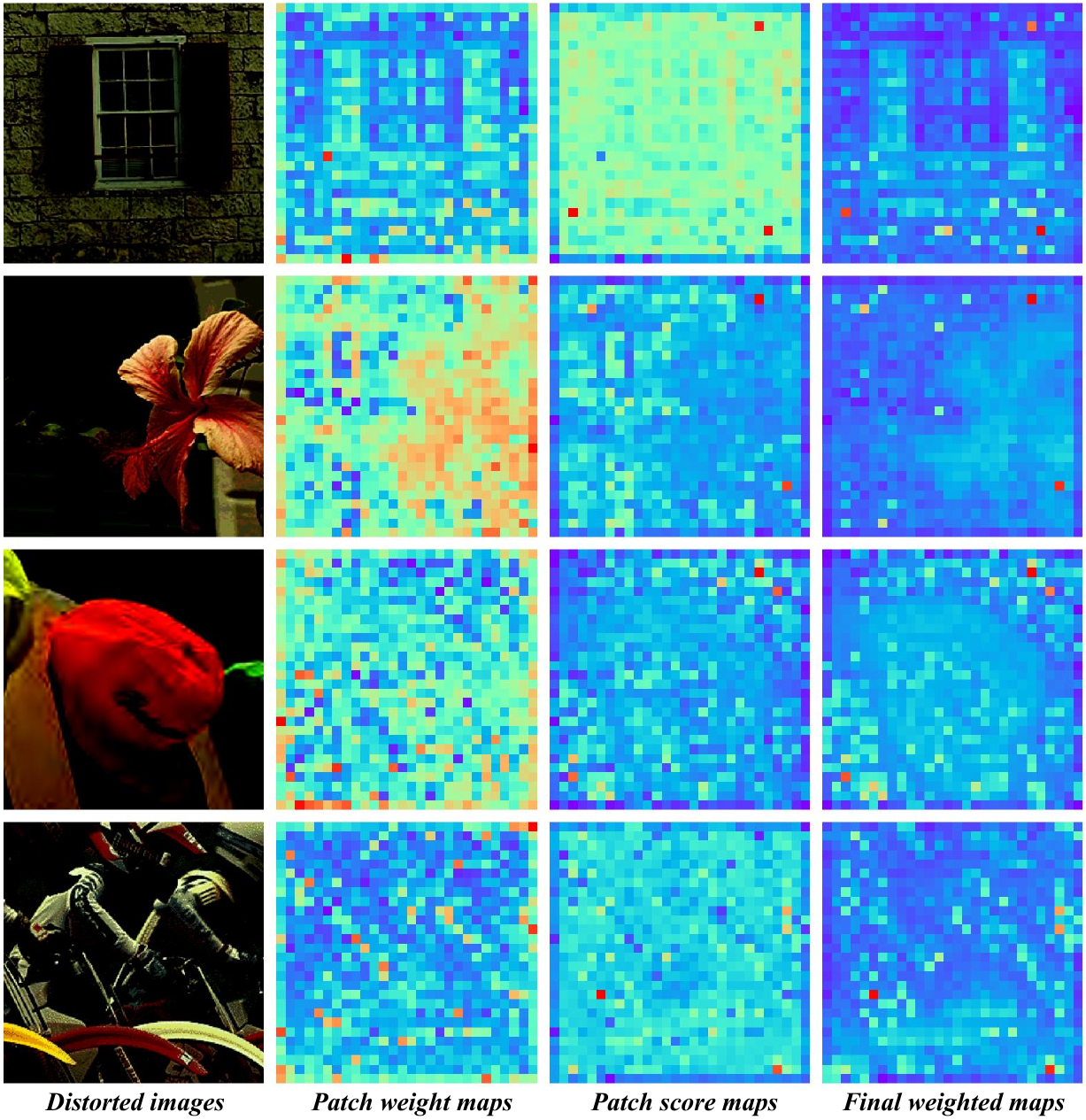}
    \caption{Visualization of weighting branch, scoring branch and final weighted maps of the test dataset of the TID2013 dataset.}
    \label{2weight}
\vspace{-3mm}
\end{figure}

\paragraph{\textbf{The visualization of the predicted weights}}
To validate the effectiveness of the two-branch structure, we generated a visual representation of the final weight distribution graph, as depicted in Fig. \ref{2weight}.
% The patch weight maps prioritize salient subjects, which are crucial areas that capture human attention. Conversely, the patch score maps focus more on regions that provide a superior visual experience.
When humans view an image, the subject itself significantly influences our perception, leading to higher weights being assigned to these regions, as illustrated in Fig. \ref{2weight}. However, high-quality regions do not always align with areas containing important semantic information.
It is evident that the two feature maps emphasize different aspects and form complementary relationships. The patch weight maps focus on the primary elements of the image, such as window frames, flowers, and the rider on the bicycle. In contrast, the patch score maps highlight finer details around the main subject, including the wall and the background behind the flower. By combining these maps, we produce a final feature weight map that not only delineates the subject's key features but also incorporates additional contextual details from the background.
% Given the disparity between notable regions and high-quality regions, the final weighted maps rely on both the patch weight and score maps.

\begin{table}[t]
  \caption{Ablation Study on Our Modules. Results Are Tested on TID2013 Dataset.}
    \centering
    \small
 
    \begin{tabular}{ccc|cc}
    \toprule
         Diffusion&  Noise embedding&  RTAB&  SRCC& PLCC\\
         \midrule
         &  &  &  0.937& 0.943\\
         \checkmark&  &  &  0.949& 0.956\\
         \checkmark&  \checkmark&  &  0.954& 0.960\\
         \checkmark&  &  \checkmark&  0.956& 0.959\\
 \checkmark& \checkmark& \checkmark& 0.962&0.966\\
 \bottomrule
    \end{tabular}
  
    \label{abl1}
\end{table}

\begin{table}[t]
  \caption{Ablation Study on Each Visual Evaluation Branch. Results Are Tested on TID2013 Dataset.}
    \centering
    \small
    \begin{tabular}{cc|cc}
    \toprule
           VCG Branch&  VDA Branch&  SRCC& PLCC\\
           \midrule
           \checkmark&  &  0.942& 0.950\\
           &  \checkmark&  0.911& 0.915\\
  \checkmark& \checkmark& 0.962&0.966\\
 \bottomrule
    \end{tabular}
  
    \label{ablbranch}
    \vspace{-3mm}
\end{table}

\begin{figure}[t]
    \centering
    \includegraphics[width=0.9\linewidth]{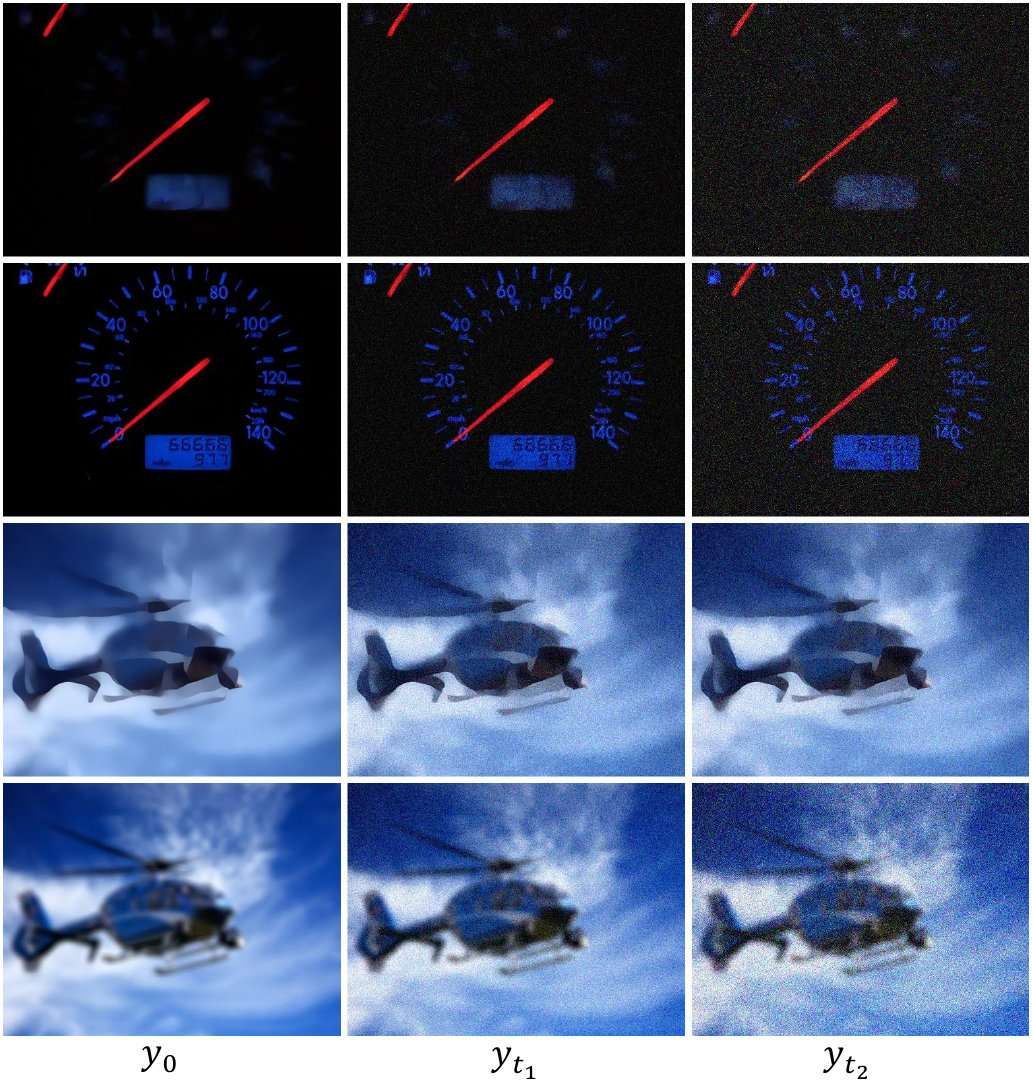}
    \caption{Visualization results of image restoration after pre-training on different datasets. Rows 1 and 3 illustrate models pre-trained on the TID2013 dataset, while rows 2 and 4 display models pre-trained on the PIPAL dataset. The images are sourced respectively from the KonIQ and Kadid datasets.}
    \label{vis_abl2}
% \vspace{-3mm}
\end{figure}

\begin{table}[t]
 \caption{The Ablation Study of Pretrained Diffusion Model-Based Image Restoration Networks on Different Dataset, ``Diff$V^2$IQA*" Denotes the Network Pre-trained on the TID2013 Dataset, While ``Diff$V^2$IQA" Indicates Pre-training on the PIPAL Dataset. }
    \centering
\small
\resizebox{0.45\textwidth}{!}{
    \begin{tabular}{c|cccccc}
    \toprule
 & \multicolumn{2}{c}{CSIQ}& \multicolumn{2}{c}{KonIQ10K}& \multicolumn{2}{c}{Kadid10K}\\
         &  SRCC&  PLCC&  SRCC&  PLCC&  SRCC& PLCC\\
     \midrule
         Diff$V^2$IQA*&  0.985&  0.989&  0.909&  0.926&  0.927& 0.929\\
         Diff$V^2$IQA&  0.986&  0.990&  0.917&  0.940&  0.935& 0.937\\
    \bottomrule
    \end{tabular}
}
   
    \label{abl2}
    \vspace{-3mm}
\end{table}

\begin{figure}[t]
    \centering
    \includegraphics[width=0.9\linewidth]{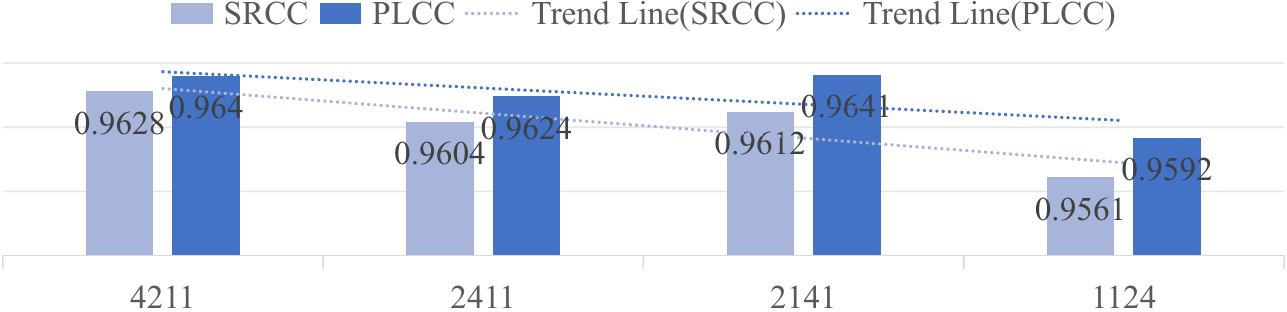}
    \caption{Ablation experiment regarding data input under various modality settings on the TID2013 dataset. For instance, the notation `4211' signifies the selection of four data blocks processed by the ViT with original data, two blocks restored using the diffusion model, and one block each from noise levels 1 and 2. }
    \label{mode_compare}
\vspace{-3mm}
\end{figure}

% \begin{table}[t]
%     \caption{\textcolor{blue}{Ablation Study of Different Scale Factors of the Image Restoration Score Branch on TID2013 Dataset. The Best Score is the Sum of SRCC and PLCC. }
% }
%     \centering
%     {\small
%     \begin{tabular}{c|ccc}
%     \toprule
%          Scale&  SRCC& PLCC &Best Score\\
%         \midrule
%             0& 0.9420& 0.9500&1.8920\\
%          0.1&  0.9596& 0.9620 &1.9216\\
%           \rowcolor{blue!5} % 整行背景色
%          0.3&  0.9612& 0.9641 &1.9253\\
%           \rowcolor{blue!5} % 整行背景色
%          0.6&  0.9592& 0.9632 &1.9224\\
%           \rowcolor{blue!5} % 整行背景色
%          1&  0.9602& 0.9651 &1.9253\\
%             \rowcolor{blue!5} % 整行背景色
%          1.5&  0.9603& 0.9642 &1.9245\\
%          2& 0.9531& 0.9579&1.9111\\
%          2.5& 0.9485& 0.9508&1.8994\\
%          4& 0.9435& 0.9509&1.8941\\
%     \midrule
%     \textit{Adaptive weight} &\textbf{0.9628}&\textbf{0.9662}& \textbf{1.9291}\\
%      \bottomrule
%     \end{tabular}
%     }
%     \vspace{-3mm}
%     \label{abl3}
% \end{table}

\begin{table*}[t]
    \caption{Ablation Study on Feature Map Selection Post-Encoding by Vision Transformer on the TID2013 Dataset.
}
    \centering
    {\small
    \begin{tabular}{c|c|cc}
    \toprule
         Mode&   Feature Map (4:2:1:1)&SRCC& PLCC \\
        \midrule
        \rowcolor{blue!5} % 整行背景色
            Continuity and Overlap&  $f_{dis}^6, f_{dis}^7 ,f_{dis}^8 ,f_{dis}^9,f_{0}^6, f_{0}^7,f_{t_1}^8, f_{t_2}^9$&0.962& 0.966\\
        \rowcolor{blue!5} % 整行背景色
         Continuity and Overlap&   $f_{dis}^1, f_{dis}^2 ,f_{dis}^3 ,f_{dis}^4,f_{0}^1, f_{0}^2,f_{t_1}^3, f_{t_2}^4$&0.956& 0.964\\
         \rowcolor{blue!5} % 整行背景色
 Continuity and Overlap& $f_{dis}^3, f_{dis}^4 ,f_{dis}^5 ,f_{dis}^6,f_{0}^3, f_{0}^4,f_{t_1}^5, f_{t_2}^6$& 0.958& 0.963\\
         Continuity and Non-overlap&   $f_{dis}^1, f_{dis}^2 ,f_{dis}^3 ,f_{dis}^4,f_{0}^6, f_{0}^7,f_{t_1}^8, f_{t_2}^9$&0.942& 0.951\\
         \rowcolor{blue!5} % 整行背景色
         Discontinuity and Overlap&   $f_{dis}^2, f_{dis}^4 ,f_{dis}^6 ,f_{dis}^8,f_{0}^2, f_{0}^4,f_{t_1}^6, f_{t_2}^8$&0.959& 0.963\\
         Discontinuity and Non-overlap&   $f_{dis}^2, f_{dis}^4 ,f_{dis}^6 ,f_{dis}^8,f_{0}^1, f_{0}^3,f_{t_1}^5, f_{t_2}^7$&0.950& 0.955\\
         Pure Randomness&   -&0.939& 0.941\\
     \bottomrule
    \end{tabular}
    }
    \vspace{-3mm}
    \label{abl4}
\end{table*}

\begin{table}[t]
    \caption{Ablation Study on Keeping Two Noised Versions of the Restored Image as Input on TID2013 Dataset.
}
    \centering
    {\small
    \begin{tabular}{ccc|cc}
    \toprule
            $y_{0}$&$y_{t_1}$&$y_{t_2}$&  SRCC& PLCC \\
        \midrule
               &&& 0.937& 0.943\\
    \checkmark&&& 0.952&0.954\\
            &\checkmark&&  0.949& 0.952\\
            &&\checkmark&  0.943& 0.951\\
            \checkmark&\checkmark&&  0.956& 0.961\\
            \checkmark&\checkmark&\checkmark&  0.962& 0.966\\
     \bottomrule
    \end{tabular}
    }

    \label{abl5}
\end{table}

\begin{table}[t]
    \caption{Ablation Study on the Selection of Query, Key, and Value in the Attention Channels of the RTAB Module on TID2013 Dataset.
}
    \centering
    {\small
    \begin{tabular}{ccc|cc}
    \toprule
         Query& Key&Value&  SRCC& PLCC \\
        \midrule
            $|x_{\text{dis}} - y_0|$& $|x_{\text{dis}} - y_{t_1}|$&$|x_{\text{dis}} - y_{t_2}|$& 0.962& 0.966\\
         $|x_{\text{dis}} - y_0|$& $|x_{\text{dis}} - y_{t_2}|$&$|x_{\text{dis}} - y_{t_1}|$&  0.958& 0.963\\
         $|x_{\text{dis}} - y_{t_2}|$& $|x_{\text{dis}} - y_{t_1}|$&$|x_{\text{dis}} - y_0|$&  0.952& 0.961\\
         \bottomrule
    \end{tabular}
    }

    \label{abl6}
\end{table}

\subsubsection{Ablation Studies}
% We systematically conducted extensive ablation experiments on our model, specifically focusing on the critical functional components. This included an in-depth analysis of the two visual evaluation branches, exploring the impact of various pre-training approaches on the diffusion repair model's performance, investigating the optimal weighting of the two visual branches when analyzed jointly, and scrutinizing the proportion of high-level image information utilized in the visual compensation guidance branch. 

\paragraph{\textbf{On the individual components' functionalities}}
In this section, we conduct experiments focusing on the individual components' functionalities, as illustrated in Table \ref{abl1}. `Diffusion' represents our diffusion restoration network, `Noise Embedding' signifies the noise level embedding strategy, while `RTAB' denotes the residual transposed attention block.
From Table \ref{abl1}, it can be observed that each component contributes significantly to the overall performance. 
The diffusion restoration model adheres to the principle of iterative denoising, generating richer high-level visual information, such as the final repaired image and the noise-containing images from the intermediate process. These images offer more detailed, clearer, and richer information to guide the network in quality evaluation.
The `Noise Embedding' strategy, meticulously crafted for encoding noise levels in distinct image states, intricately refines the model's capability to navigate diverse image conditions. This strategy empowers the network to discern to a certain extent between crucial original distorted image details and additional introduced high-level visual information, facilitating more effective learning of feature information. 
The `RTAB’ module, as the core architecture of the VDA branch, emulates the FR-IQA method by incorporating the `reference' for image evaluation. It focuses on analyzing the differences between high-level visual images and input images. Coupled with an attention mechanism, the module enhances the utilization of the overall augmented features, thereby enhancing overall performance.

\paragraph{\textbf{On the individual visual branch' functionalities}}
Ablation experiments are conducted for each functional branch, as detailed in Table \ref{ablbranch}. The optimal results are achieved when both branches collaborated in the quality evaluation, with a notable performance gap observed between the VDA and VCG branches.
The experiments confirm that the ViT architecture provides superior feature capture capabilities compared to ResNet. Moreover, both branches prove effective, with the VCG mechanism demonstrating greater robustness than the VDA mechanism. Despite this, the two branches function through distinct processes: the VCG branch employs high-level visual information to guide the network's evaluation of the original image in a conventional NR-IQA manner, whereas the VDA branch analyzes discrepancies between high-level visual information and the original image, akin to an `FR-IQA' approach. Consequently, although the VDA branch alone may not produce outstanding results, the integration of both branches significantly improves performance by combining their complementary perspectives, offering a more holistic and robust evaluation.
% This disparity arises primarily because the VDA branch has a smaller overall number of parameters and utilizes a ResNet-based architecture, which is less effective at feature extraction compared to ViT. 
% The VDA branch serves as a mechanism to mitigate the risk of overfitting in the VCG branch. Additionally, it maximizes the utilization and exploitation of feature information from the enhanced data from the diffusion restoration model. 
% This observation underscores the significance of the intrinsic features of the distorted image in quality assessment and further affirms the superiority of the transformer architecture over the ResNet architecture in capturing detailed features. 

\paragraph{\textbf{On the variations in performance across different pre-training datasets for the diffusion model}}
We additionally perform extended experiments on diffusion restoration networks, specifically investigating the influence of the training effectiveness of the restoration networks on the overall performance of the model. 
We pre-train our diffusion restoration network on two datasets: the TID2013 dataset and the PIPAL dataset. Each dataset utilize all available samples for training, with consistent parameters except for the dataset source. Following the initial training, we conduct quantitative experimental tests on three additional datasets, comprising two synthetic distortion datasets and one authentic distortion dataset, as illustrated in Table \ref{abl2}. 
The PIPAL dataset encompasses a more extensive training dataset (PIPAL: 23,200, TID2013: 3,000) and a greater variety of image distortion categories (PIPAL: 116, TID2013: 24) compared to the TID2013 dataset. Consequently, image restoration networks with enhanced generalization performance can be trained. From Table \ref{abl2}, it is evident that when dealing with a dataset like CSIQ, which is not particularly complex, the difference in results between the two datasets is not significant. However, when faced with a larger synthetic distortion dataset or a authentic distortion dataset, the diffusion restoration network trained on PIPAL exhibits superior performance. 

Additionally, we conduct visualizations of repaired images along with noise-containing images acquired under different pre-training datasets, as depicted in Fig. \ref{vis_abl2}. It is evident that the enhanced images obtained through training on the larger and more diverse PIPAL dataset are clearer and align more closely with the standard of high-quality images. 
In contrast, under TID2013 training, the network faces challenges in recovering and enhancing certain detailed texture information. 

\paragraph{\textbf{On the context of employing different ratio schemes for original, augmented, and noisy images}}
We conduct noise level embedding in the ViT-based assessment branch for the original image, the enhanced image, and the noise-containing images. Subsequently, feature information of varying lengths from different images is selected for feature splicing, facilitating network feature recognition. To validate the rationality of our chosen ratio, we designed experiments, as illustrated in Fig. \ref{mode_compare}. 
% We selected a total of 8 feature maps for feature splicing. The notation `4211' in the figure signifies that 4, 2, 1, and 1 blocks of feature maps were chosen for splicing in the original image, the enhanced image, the image with noise level 1, and the image with noise level 2, respectively.
As observed in Fig. \ref{mode_compare},  similar to the earlier experimental findings, consistent conclusions emerge: a greater emphasis on the feature information of the original image is necessary. Overemphasizing the additional information introduced by the image enhancement network can prove detrimental to the network's learning process. 
% In summary, our selection of feature maps is both rational and effective, enabling the network to more effectively extract data feature information and yield superior results. 

\paragraph{\textbf{On the feature map selection post-encoding by vision transformer}}
In the VCG branch, the input images are initially encoded using ViT, resulting in the generation of nine corresponding feature maps for each image. In order to select the most appropriate feature maps for use, we conduct ablation studies, as shown in Table \ref{abl4}. 
The Continuity versus Discontinuity indicator determines whether the selected feature maps are positionally continuous. The Overlap versus Non-overlap indicator determines whether the selected feature maps overlap in position between the original distorted image and the diffusion-restored image. In the case of pure randomness, the feature maps are selected at random each time, while the overall ratio of 4:2:1:1 is maintained. 
As demonstrated in Table \ref{abl4}, the principle of overlap during feature selection has a significant impact on the results. The results highlighted in blue in the table demonstrate that optimal performance can be achieved when different feature maps are selected, regardless of continuity, as long as overlap is maintained. When feature maps are non-overlapping, the higher-level information introduced by the distorted image and the diffusion-repaired image is distributed across different feature layers, preventing effective higher-level visual guidance. 
When the selection of feature maps is entirely random, the utilisation of feature data is increased. However, the difficulty in matching distorted image features with higher-level visual feature information prevents the formation of an effective visual feature guide, resulting in suboptimal performance.

\paragraph{\textbf{On keeping two noised versions of the restored image}}

Inspired by the use of multi-level and multi-scale feature representations in models such as VCRNet and HyperIQA, we leverage data from different stages, time steps, and feature distributions within the denoising process of the diffusion model as high-level visual features. To assess the effectiveness of these feature selections, ablation experiments are conducted, with results presented in Table \ref{abl5}.
The results indicate that the optimal performance is achieved when both the final reconstructed image and two noise-containing images are used as inputs. This configuration enriches the model with diverse high-level visual information and learnable relationships across different time steps, thereby improving the overall network performance.

\paragraph{\textbf{On the selection of query, key, and value in the attention channels of the RTAB module}}

In the RTAB module, we design the attention mechanism based on the difference between the enhanced images and the original distorted image, as shown in Eq. \ref{attn}. To evaluate the impact of using this difference as query, key, and value, we conduct an ablation experiment, as presented in Table \ref{abl6}. 
It can be observed that better results are achieved when the difference between the distorted image and the one with more noise is used as the Value ($|x_{\text{dis}} - y_0|$). In the VCG branch, we prioritize features from the final recovered image to focus on the end recovery result. Conversely, in the VDA branch, which complements the VCG branch, greater emphasis is placed on the noise-containing images. This approach allows for a more comprehensive utilization of high-level visual features and helps mitigate the risk of overfitting in the VCG branch, thereby improving overall accuracy.

% \begin{table}[t]
%     \centering
%     \small
% \caption{Computational Complexity of Each NR-IQA Method }
% \label{time}
%     \begin{tabular}{ccc}
%     \toprule
%          Method&  Time (sec.)& Params (MB)\\
%          \midrule
%          CNN\cite{kang2014convolutional}&  0.07& 2.77\\
%          DeepIQA\cite{bosse2017deep}&  0.12& 39.97\\
%          TSCNN\cite{yan2018two}&  0.19& 5.49\\
%  DBCNN\cite{zhang2018blind}& 0.09&58.41\\
%  RAN4IQA\cite{ren2018ran4iqa}& 0.73&158.14\\
%  Hall-IQA\cite{lin2018hallucinated}& 1.36&709.10\\
%  VCRNet\cite{pan2022vcrnet}& 0.62&473.56\\
%          Diff$V^2$IQA&  9.65& 277.71\\
%          \bottomrule
%     \end{tabular}

% \end{table}

% \subsubsection{Computational Complexity}
% To evaluate the computational complexity of each NR-IQA model, we utilize the average GPU testing time per image and the number of parameters. The results are presented in Table \ref{time}. 
% It is evident that the number of parameters in our model is moderate, totaling only 277.71 MB, which is less than both VCRNet and Hall-IQA. However, in terms of inference time, our model exhibits a prolonged duration due to the iterative denoising process involving the diffusion model. This extended inference time results from the iterative mechanism inherent in the diffusion model, presenting a significant factor affecting the efficiency of our model. This limitation underscores the need for future improvements and further research to enhance the model's performance. 

\section{Limitation and Future Work}
Firstly, the model must undergo pre-training of the diffusion restoration network, a process that is not as succinct as certain end-to-end direct quality evaluation networks. Furthermore, the diffusion restoration network exhibits a certain degree of dependence on the pre-trained dataset. Additionally, owing to the iterative denoising principle inherent in the diffusion model, there is an inherent extension of inference time to some extent. 
Finally, our model's performance on authentic distortion datasets needs further improvement. We believe progress can be made in the following ways:
\begin{itemize}
\item \textbf{Better Datasets:} The development of more realistic simulated datasets could enable the model to learn generalizable restoration features that are closer to authentic distortions. Alternatively, the availability of labeled authentic distortion datasets could enhance model training.
\item \textbf{Stronger Image Restoration Models:} The emergence of models with superior image enhancement capabilities could lead to better restoration of authentic distorted images, providing higher-level visual guidance.
\item \textbf{Effective Data Augmentation Strategies:} For unlabeled authentic distortion datasets, data augmentation could be employed to create new data pairs for model training. For instance, in Re-IQA, the authors designed 25 types of distortions to augment authentic distortion datasets, creating new data pairs for contrastive learning. This approach is a promising avenue for further exploration.
\end{itemize}

In our future work, we will concentrate on alleviating these challenges and intend to progressively transition our research towards more advanced aspects of video tasks.

\section{Conclusion}
% \textbf{\textit{Contribution:}}
In this paper, we propose a new Diff$V^2$IQA model specifically designed for NR-IQA tasks. Notably, we pioneer the incorporation of the diffusion model into the NR-IQA domain. 
Followed free energy principle, we have innovatively designed a novel diffusion restoration network with heightened capability in capturing specific and intricate features. Furthermore, to comprehensively analyze the extracted high-level visual information, we devised two distinct visual evaluation branches. The first is a visual compensation guidance branch built on the transformer architecture, integrating a noise embedding strategy. The second involves a visual difference analysis branch rooted in the ResNet architecture and we proposed RTAB to better catch the detailed difference between distorted and restored images. These two branches assess quality from distinct perspectives and ultimately converge their analyses to derive the overall quality score. 
Extensive experiments substantiate the effectiveness of our model, showcasing its superiority over SOTA methods in recent years. Rigorous experimentation on various aspects of our network design and the selection of hyperparameters further attests to the rationality and effectiveness of our proposed approach. In conclusion, our Diff$V^2$IQA model emerges as highly effective, achieving SOTA performance in the domain of NR-IQA.

\bibliographystyle{IEEEtran}
\bibliography{reference}

\end{document}